\documentclass[11pt,journal,letterpaper,twoside,onecolumn]{IEEEtran}

\usepackage{amsmath, amssymb, colortbl, enumerate,url}
\usepackage[final]{graphicx}
\usepackage[latin1]{inputenc}
\usepackage{multirow,array}

\newcommand{\mB}{\mathbf{B}}
\newcommand{\vb}{\mathbf{b}}
\newcommand{\mI}{{\mathbf I}}
\newcommand{\vg}{\mathbf{g}}

\newcommand{\vs}{\mathbf{s}}

\newcommand{\ve}{\mathbf{e}}

\newcommand{\vx}{\mathbf{x}}

\newcommand{\vxh}{\hat{\mathbf{x}}}

\newcommand{\vy}{\mathbf{y}}

\newcommand{\vz}{\mathbf{z}}

\newcommand{\vp}{\mathbf{p}}

\newcommand{\vk}{\mathbf{k}}

\newcommand{\vn}{\mathbf{n}}

\newcommand{\vf}{\mathbf{f}}

\newcommand{\mS}{\mathbf{S}}

\newcommand{\mG}{\mathbf{G}}

\newcommand{\mD}{\mathbf{D}}

\newcommand{\mM}{\mathbf{M}}

\newcommand{\mX}{\mathbf{X}}

\newcommand{\mT}{\mathbf{T}}

\newcommand{\mF}{\mathbf{F}}

\newcommand{\mU}{\mathbf{U}}

\newcommand{\mH}{\mathbf{H}}

\newcommand{\eps}{\boldsymbol{\epsilon}}

\newcommand{\vxi}{\boldsymbol{\xi}}

\newcommand{\mW}{\mathbf{W}}

\usepackage[lined,linesnumbered,ruled]{algorithm2e}

\newcommand\scalemath[2]{\scalebox{#1}{\mbox{\ensuremath{\displaystyle #2}}}}

\usepackage{etoolbox}
\makeatletter
\patchcmd{\@makecaption}  {\\}  {.\ }  {}  {}

\begin{document}
\title{Robust Online Video Super-Resolution Using an Efficient Alternating Projections Scheme}

\author{Ricardo~Augusto~Borsoi
\thanks{R.A. Borsoi is with the Department
of Electrical Engineering, Federal University of Santa Catarina, Florian\'opolis, SC, Brazil, and with the Lagrange Laboratory, Universit\'e C\^ote D'Azur, CNRS, Nice, France. e-mail: raborsoi@gmail.com.}
}

\maketitle

\begin{abstract}
Video super-resolution reconstruction (SRR) algorithms attempt to reconstruct high-resolution (HR) video sequences from low-resolution observations. Although recent progress in video SRR has significantly improved the quality of the reconstructed HR sequences, it remains challenging to design SRR algorithms that achieve good quality and robustness at a small computational complexity, being thus suitable for online applications. In this paper, we propose a new adaptive video SRR algorithm that achieves state-of-the-art performance at a very small computational cost. Using a nonlinear cost function constructed considering characteristics of typical innovation outliers in natural image sequences and an edge-preserving regularization strategy, we achieve state-of-the-art reconstructed image quality and robustness. This cost function is optimized using a specific alternating projections strategy over non-convex sets that is able to converge in a very few iterations. An accurate and very efficient approximation for the projection operations is also obtained using tools from multidimensional multirate signal processing. This solves the slow convergence issue of stochastic gradient-based methods while keeping a small computational complexity. Simulation results with both synthetic and real image sequences show that the performance of the proposed algorithm is similar or better than state-of-the-art SRR algorithms, while requiring only a small fraction of their \mbox{computational cost.}
\end{abstract}


\section{Introduction}

Super-resolution reconstruction (SRR) is a technique that aims to obtain high-resolution (HR) images of a given scene from low-resolution (LR) observations~\cite{Park03,Nasrollahi14_full,yue2016SRR_review__SP}.
Due to its ability to transcend the physical limitations of conventional imaging sensors, SRR finds applications in different areas ranging from standard end-user digital cameras to forensics~\cite{villena2018SRR_forensics}, remote sensing~\cite{borsoi2018FuVar} and medical imaging~\cite{borsoi2018SRR_EIT}.

SRR methods can be divided in image and video SRR algorithms.
While in image SRR only one HR image is reconstructed from either a single or multiple LR observations, video SRR methods aim to reconstruct an entire HR image sequence.
Besides improving the spatial resolution of the LR image sequence, video SRR methods must also ensure the consistency of the reconstructed HR frames over time. This is usually performed by introducing information about the high correlation between adjacent frames in the form of a \mbox{temporal regularization~\cite{Borman99,Zibetti07,Belekos10_full}.}


Although a significant number of image and video SRR algorithms have already been proposed, recent advances have been mostly focused at improving the quality of the reconstructed images.
This has been the case, for instance, in 
non-parametric spatial kernel regression methods~\cite{Takeda09}, variational Bayesian methods \cite{babacan2011variationalBayesianSRR,liu2014bayesianVideoSRR}, non-local methods~\cite{Protter09,barzigar2016videoSRRscobep,chang2018singleSRR_nonLocal__SP,hu2018noiseRobustMultiscalePyramid__SP}, and more recently in deep-learning-based methods \cite{liu2017robustSRRneuralNets,kappeler2016videoSRRneuralNets,tao2017detailRevealingSRRneuralNets,wang2019generativeSingleImageSRR__SP}.
Although these techniques led to considerable improvements in the quality of the reconstructed images in state-of-the-art SRR algorithms, the computational complexity associated with them is very high, making them unsuitable for real-time applications.
Furthermore, deep-learning methods, while possibly faster to deploy, require large amounts of training data and extensive training procedures, which must also be repeated whenever the test conditions change, otherwise the SRR performance may degrade significantly~\cite{bhowmik2017trainingFreeCNN}.

Thus, even though many recent SRR methodologies achieve good reconstruction results, real-time applications require low-complexity algorithms.
Among the simpler video SRR algorithms, the regularized least mean squares (R-LMS)~\cite{Elad99,Elad99pami} stands out due to its extreme simplicity and good performance. 
Using a stochastic gradient solution to optimize a quadratic cost function, the complexity of the R-LMS is small even when compared to other low-complexity algorithms that were proposed for global translational motion models, such as the median estimator in~\cite{farsiu2004fastRobustSRR} and the adaptive Wiener filter of~\cite{Hardie07}. 
Furthermore, the R-LMS algorithm's behavior has also been characterized mathematically~\cite{Costa07}, what led to principled parameter design methodologies~\cite{Costa08}.

However, although the R-LMS has an SRR image quality that is comparable to that of more elaborate algorithms under simple motion models and is naturally robust to additive noise and registration errors~\cite{Costa09regErrors}, its performance degrades considerably in the presence of innovation outliers caused by occlusions due to moving objects and sudden scene changes~\cite{borsoi2019SRR_adap_TIP}.

%
This lack of robustness has been recently addressed by studying the proximal point cost function representation of the R-LMS iterative equation. The R-LMS performance degradation was shown to be caused by its slow convergence when faced with (statistically) typical innovation outliers encountered in natural image sequences. This resulted in the proposal of a new cost function and a novel stochastic gradient-based algorithm (named LTSR-LMS) which achieved a significant performance improvement at a comparable computational complexity~\cite{borsoi2019SRR_adap_TIP,borsoi2017newSRRrobustnessInn}.
Nonetheless, the stochastic gradient nature of this (LTSR-LMS) algorithm meant that it still lacked robustness when compared to more costly and elaborate algorithms~\cite{borsoi2019SRR_adap_TIP}. Furthermore, the quadratic nature of the R-LMS and LTSR-LMS cost functions leads to their inability to adequately preserve image edges, resulting in either blurry or noisy reconstructions.
Thus, it is of necessary to develop algorithms that can achieve better robustness and edge preservation at a similar computational complexity, closing the gap between low-complexity algorithms and state-of-the-art performance.

In this paper, a new adaptive video SRR algorithm with improved robustness to noise and innovation outliers is proposed. The algorithm derivation is divided in two main contributions/parts.
First, the solution to the (quadratic) cost function of the LTSR-LMS super-resolution algorithm~\cite{borsoi2019SRR_adap_TIP} is interpreted as an approximate inversion of a given multidimensional multirate system using a gradient descent algorithm. Thus, using tools from multirate system theory, we abandon the gradient-based approach and instead seek for a more accurate approximate solution that can be computed \textit{a priori} and thus applied much more efficiently. This significantly improves robustness by addressing the convergence issue of the stochastic gradient descent in previous algorithms while still retaining the quality and robustness of the cost function~\cite{Elad99pami,Elad99,borsoi2019SRR_adap_TIP}.

Secondly, we consider a nonlinear Wavelet-based edge-preserving regularization strategy to improve the robustness to additive noise, allowing for smoother reconstructed image sequences without compromising sharp image edges.
%
In order to solve the resulting nonlinear cost function at each time instant, we propose a specific alternating projections scheme over non-convex sets that converges to a good solution in a very few iterations. Furthermore, efficient solutions to the projection operations are derived by exploring multirate system theory and the orthogonality property of the Wavelet transform. This allows for a computational complexity still comparable to that of the gradient-based solutions, albeit with a significantly improved quality.
The proposed algorithm is shown to outperform state-of-the-art video SRR algorithms with only a small fraction of their computational cost.

This paper is organized as follows. The imaging model is defined in Section~\ref{sec:notation}. In Section~\ref{sec:TSR_LMS}, the image and video SRR problem is presented, as well as the normal equations to the LTSR-LMS cost function~\cite{borsoi2019SRR_adap_TIP}. In Section~\ref{sec:multirate_srr_cf}, the connection between the video SRR problem and multirate systems is addressed. The basic mathematical theory of multidimensional multirate signal processing is presented in Section~\ref{sec:multirate_theory}. In Section~\ref{sec:edge_preserv}, a new efficient edge-preserving SRR algorithm using an alternating projection scheme is proposed. Computer simulations are performed to assess the performance of the algorithms in Section~\ref{sec:results}. Finally, the conclusions are presented in Section~\ref{sec:concl}. The main symbols and notation used in this paper are presented in Table~\ref{tab:notations}.


\begin{table}[ht]
\centering
\renewcommand{\arraystretch}{1.2}
\caption{List of Notation and Symbols}
\label{tab:notations}
\begin{minipage}{0.45\linewidth}
\resizebox{0.95\width}{!}{%
\begin{tabular}{ll}
    \multicolumn{2}{c}{\emph{Scalars}} \\
    $M$, $N$ & Dimensions of the HR and LR images \\
    $L$ & Number of filters/channels in a filterbank \\
    $Q$ & Number of decomposition levels of \\& the Wavelet transform \\
    $J$ & Number of alternating projections iterations \\
    $R$ & Number of sets forming the constraints \\
    $d$ & downsampling rate ($M/N$) \\
    $k$ & Discrete time instant \\
    %
    %
    $\alpha_{(\cdot)}$, $\lambda_{(\cdot)}$ & Regularization parameters \\
    %
    $\sigma$, $\gamma$, $\tau$ & Parameters defining the sets $\Omega_1$ and $\Omega_2$ \\
    \multicolumn{2}{c}{\emph{Vectors}} \\
    $\vx(k)$, $\vy(k)$   &   Vectorized versions of  $\mX(k)$ and $\mathbf{Y}(k)$ \\
    $\vs(k)$, $\ve(k)$ & Innovations and sensor noise \\
    $\vxh_j(k)$ & Estimated HR image at the $j^{\rm th}$ iteration  \\
    $\vn$ & Discrete spatial index/position \\
    $\vk_{\ell}^{\text{SRR}}$ & 2D vectors, for $\ell=1,\ldots,d^2$, belonging \\& to the set $\{1,\ldots,d\}\times\{1,\ldots,d\}$ \\
    \multicolumn{2}{c}{\emph{Matrices}} \\
    $\mX(k)$,$\mathbf{Y}(k)$   &   HR and LR images, represented as matrices \\
    $\mG(k)$ & Matrix representing the motion between \\& the HR images at instants $k-1$ and $k$ \\
    $\mM$ & Matrix representing decimation in the spatial \\& domain as $\vn\mapsto\mM\vn$, where $|\det(\mM)|=d^2$ \\
    $\mW$ & Discrete Wavelet transform matrix \\
    $\mH$, $\mS$  & Matrices representing the 2D convolution of \\& the vectorized HR image by $h(\vn)$ and $s(\vn)$ \\& in the form of matrix-vector products \\
    $\mD$ & Downsampling matrix \\
    %
    %
    %
    %
    
\end{tabular}
}
\end{minipage}
\begin{minipage}{0.54\linewidth}
\resizebox{0.95\width}{!}{%
\begin{tabular}{ll}
    \multicolumn{2}{c}{\emph{2D Discrete Signals}} \\
    $x(\vn)$ & HR image represented as a 2D discrete signal \\ 
    $h(\vn)$, $s(\vn)$  & 2D discrete signals representing sensor blur \\& and a high pass (Laplacian) filter \\
    $\delta(\vn)$ & 2D discrete Kronecker delta function \\
    $f_{\ell}(\vn)$, $g_{\ell}(\vn)$  &  $\ell^{\rm th}$ analysis and synthesis filters in a filterbank \\
    $f_{\ell}'(\vn)$, $g_{\ell}'(\vn)$  &  $\ell^{\rm th}$ analysis and synthesis filters in the \\& approximate inverse filterbank \\
    \multicolumn{2}{c}{\emph{Z-Domain Variables}} \\
    $X(z)$,$X_{\vk_i}(z)$ &  Z-transform of $x(\vn)$ and its $i^{\rm th}$ polyphase component \\
    $F_{\ell}(z)$,$G_{\ell}(z)$ & 2D Z-transform of $f_{\ell}(\vn)$ and $g_{\ell}(\vn)$ \\
    $F_{\ell,\vk_i}(z)$,$G_{\ell,\vk_i}(z)$ & $i^{\rm th}$ polyphase component of $f_{\ell}(\vn)$ and $g_{\ell}(\vn)$ \\
    %
    %
    $\vx_p(z)$ &  Vector containing all polyphase components of $X(z)$ \\
    $\vf_{p,\ell}(z)$, $\vg_{p,\ell}(z)$ & Vector containing all polyphase components \\& of the $\ell^{\rm th}$ analysis and synthesis polyphase filters \\
    $\mF_p(z)$, $\mG_p(z)$ & Matrices combining the polyphase components \\& of all $L$ filters in a filterbank \\
    $\mT(z)$, $\mU(z)$ & Polyphase transfer matrix and its inverse \\
    \multicolumn{2}{c}{\emph{Sets, Norms, Operators and Functions}} \\
    $*$ & 2D convolution operator \\
    $\|\cdot\|_p$   &   Euclidean $p$-norm, $\|\vb\|_p=\big(\sum_i |b_i|^p\big)^{1/p}$ \\
    $\|\cdot\|_F$   &   Frobenius norm, $\|\mB\|_F=\big(\sum_i\sum_j |B_{i,j}|^2\big)^{1/2}$ \\
    %
    $\|\cdot\|_{\mathcal{L}}$   & Unit sphere seminorm, $\|f\|_{\mathcal{L}}=\int_{\|z\|=1} |f(z)|^2dz$ \\
    $\mathcal{R}(\cdot)$ & Spatial regularization function \\
    $\mathcal{N}(\mM)$ & Fundamental parallelepiped of $\mM$ \\
    %
    %
    %
    $\operatorname{thr}_p(\cdot,\lambda_{\tau})$ & Soft ($p=1$) or hard ($p=0$) thresholding operator \\
    $\Omega_r$ & Constraint sets in the alternating projections scheme \\
    $\mathcal{P}_r(\cdot)$ & Projection operator onto the set $\Omega_r$ \\
    $\Lambda$ & Set of parameters used in the alternating projections \\ 
    %
    %
\end{tabular}
}
\end{minipage}
\end{table}






\section{Image Acquisition Model}
\label{sec:notation}

Considering the matrix representation of an observed LR image $\mathbf{Y}(k)$ of size $N \times N$ and the matrix representation of the original HR digital image $\mathbf{X}(k)$, of size $M \times M$ ($M > N$), the acquisition process can be modeled as~\cite{Park03}
\begin{equation}\label{eq:aquis}
    \vy(k) = \mD\mH\vx(k) + \ve(k),
\end{equation}
where $N^2\times1$ vector $\vy(k)$ and $M^2\times1$ vector $\vx(k)$ are the vectorizations of the degraded and original images $\mathbf{Y}(k)$ and $\mathbf{X}(k)$, respectively, at discrete time instant $k\in\mathbb{Z}_+$. Matrix $\mD$, of size $N^2 \times M^2$, is the decimation matrix and models the subsampling taking place in the sensor. Matrix $\mH$, of size $M^2 \times M^2$, models the blurring taking place in the acquisition process, which is assumed to be known \textit{a priori} and shift invariant.
The product $\mH\vx(k)$ is assumed to be equivalent to the vectorization of the bidimensional (spatial) convolution of the HR image $\mX(k)$ by a discrete convolution mask $h(\vn)$, where $\vn\in\mathbb{Z}^2$ denotes the discrete spatial position.
Vector $\ve(k)$, of the same size as $\vy(k)$, models the additive observation noise, whose properties can be determined from camera tests. The dynamics of the input signal is modeled by
\begin{equation}\label{eq:dinam}
    \vx(k) = \mG(k) \vx(k-1) + \vs(k),
\end{equation}
where $\mG(k)$ is a matrix that describes the relative displacement between the HR images in time~$k-1$ and in time~$k$. Vector $\vs(k)$ models the innovations in the HR image sequence, which are usually caused by scene changes such as occlusions.

\section{Adaptive Video Super-Resolution}
\label{sec:TSR_LMS}

A large number of SRR algorithms are based on the minimization of the reconstruction error~\cite{Park03,Nasrollahi14_full}
\begin{equation} \label{eq:epsilon}
    \eps(k) {}={} \vy(k)-\mD\mH \vxh(k),
\end{equation}
where $\vxh(k)$ is the estimated HR image, and $\eps(k)$ can be interpreted as the estimate of $\ve(k)$ in~\eqref{eq:aquis}. 
Since images considered in most practical applications are intrinsically smooth~\cite{ruderman1994statisticsNaturalImages}, this \textit{a priori} knowledge can be added to the estimation problem in the form of a regularization by constraining the solution that minimizes~$\|\eps(k)\|^2$, resulting in the following cost function
\begin{equation} \label{eq:Lagrangian}
    \mathcal{L}_{\text{R}}(k) = \| \vy(k)-\mD\mH \vxh(k) \|^{2}_2
    + \alpha \mathcal{R}(\vxh(k)),
\end{equation}
where function $\mathcal{R}(\cdot):\mathbb{R}^M\to\mathbb{R}$ is a spatial regularization term and $\alpha\in\mathbb{R}_+$ is a parameter that balances the contribution of the terms in the cost function.
The function $\mathcal{R}$ can either be a quadratic (Thikonov) regularization, or it can comprise more advanced edge-preserving regularization schemes which penalize the Total Variation~\cite{farsiu2004fastRobustSRR,liu2014bayesianVideoSRR,kohler2016robustSRRirsw} of the image or the L$_p$-norm ($0\leq p\leq1$) of its representation in some sparsifying basis such as the Wavelet domain~\cite{figueiredo2003EM_waveletRestoration,figueiredo2007MM_waveletRestoration,afonso2010fastImgRecVariableSplitting,matakos2013acceleratedRestoration,buades2005reviewDenoisingNLM}.
Note that the performance surface in~\eqref{eq:Lagrangian} is defined for each discrete time instant~$k$.

Although edge-preserving regularization terms usually lead to reconstructed images with better perceptual quality, the minimization of the resulting cost function requires expensive, iterative optimization procedures~\cite{figueiredo2003EM_waveletRestoration,daubechies2004sparse_ISTA,beck2009sparse_FISTA,beck2009denoisingDeb_FISTA,matakos2013acceleratedRestoration,ng2010TV_admm,boyd2011ADMM,afonso2010fastImgRecVariableSplitting,goldstein2009splitBregmanL1,afonso2011ADMM_imagingInvProb}.
This leads to a computational cost that is generally incompatible with the requirements of real-time processing applications. Thus, to achieve a faster execution times, adaptive SRR algorithms usually employ the quadratic Thikonov regularization
\begin{align} \label{eq:thikovov_sp_reg}
    \mathcal{R}(\vxh(k)) {}={} \|\mS\vxh(k)\|^2_2,
\end{align}
where matrix $\mS$ is the Laplacian operator. The product $\mS\vxh(k)$ is also assumed to be equivalent to the vectorization of the bidimensional (spatial) convolution of the HR image estimate $\hat{\mX}(k)$ by a discrete convolution mask $s(\vn)$, where $\vn\in\mathbb{Z}^2$.
%
%
%
In the remainder of this section, we will review the video SRR problem as previously considered in \cite{Elad99,borsoi2017newSRRrobustnessInn,borsoi2019SRR_adap_TIP}, which used a Thikonov spatial regularization as defined in~\eqref{eq:thikovov_sp_reg}.
This problem will then be extended to consider an edge-preserving spatial regularization for the proposed solution in  Section~\ref{sec:edge_preserv}.

Adaptive video SRR algorithms attempt to estimate $\vx(k)$ for each time instant $k\in\mathbb{Z}_+$ using information from previously estimated images contained in~$\vxh(k-1)$.
For instance, the regularized Least Mean Squares (R-LMS) algorithm attempts to do so by minimizing a stochastic version of~\eqref{eq:Lagrangian} using an iterative gradient descent method, where the temporal information is included in the solution by means of the initialization of the optimization process (\textit{i.e.} the solution $\vxh(k)$ in the first iteration is initialized as $\mG(k)\vxh(k-1)$)~\cite{Costa09regErrors,Elad99pami,Elad99}.

More recent approaches proceeded to remove the dependency on the initialization and improve the robustness to innovation outliers (\textit{i.e.} sudden content changes between frames) by introducing an additional term which preserves (only) the estimated details across time, allowing the resulting algorithm to adapt faster to represent new information observed in~$\vy(k)$~\cite{borsoi2019SRR_adap_TIP,borsoi2017newSRRrobustnessInn}. This results in the following cost function~\cite{borsoi2019SRR_adap_TIP}:
\begin{align} \label{eq:Lagrangian2}
    \mathcal{L}_{\text{T}}(k) {}={} & \| \vy(k)-\mD\mH \vxh(k) \|^{2}_2
	+ \alpha \|\mS\vxh(k)\|^2_2
	\nonumber\\ &
	+ \alpha_{\tiny{\text{T}}} \big\|\mS\big[\vxh(k) -\mG(k)\vxh(k-1)\big]\big\|^2_2.
\end{align}
The optimization of the cost function $\mathcal{L}_{\text{T}}(k)$ is tantamount to solving the following linear system of equations:
\begin{align} \label{eq:sys_eq_trlms}
    & \big[\mH^\top\mD^\top\mD\mH 
    + (\alpha+\alpha_{\tiny{\text{T}}}) \mS^\top\mS \big] \vxh(k) {}={}
    \mH^\top \mD^\top \vy(k) 
    \nonumber \\
    & \hspace{4.2cm} + \alpha_{\tiny{\text{T}}} \mS^\top\mS\mG(k)\vxh(k-1).
\end{align}
However, directly solving \eqref{eq:sys_eq_trlms} requires a computational cost which is cubic in the number of HR pixels, and is thus impractical for reasonable image sizes.
As opposed to the image deblurring case where (under appropriate boundary conditions) only a block circulant matrix must be inverted, which can be done efficiently, the matrix in the right hand side of~\eqref{eq:sys_eq_trlms} is only block Toeplitz, which significantly complicates the solution.

Some methods attempt to deal with this problem by using preconditioning techniques and conjugate gradient descent methods~\cite{nguyen2001preconditioningSRR,pelletier2012preconditioningSRRedge}.
However, the computational cost of these methods is still not compatible with online processing requirement in many devices (\textit{e.g.} when operating on a power budget).

Approaches to provide a very low-cost solution attempt to use a gradient descent method to optimize~\eqref{eq:Lagrangian2} \cite{borsoi2017newSRRrobustnessInn,borsoi2019SRR_adap_TIP}. Although these methods reach very good quality for well-behaved sequences, they are still significantly less robust to innovation outliers when compared to an exact solution to~\eqref{eq:sys_eq_trlms}. This is caused due to the slow convergence of iterative methods, since the linear system in~\eqref{eq:sys_eq_trlms} is very ill-conditioned.
%


Although preconditioning techniques could be applied to alleviate this issue, they can be difficult to design (especially if edge preserving regularization terms are considered), and the resulting algorithm still presents a significant computational complexity. Instead of adopting iterative solutions, we will pursue a different approach by studying a representation of~\eqref{eq:sys_eq_trlms} as a multidimensional multirate system. This will allow us to obtain an approximate closed form solution to~\eqref{eq:sys_eq_trlms} that can be computed very efficiently, avoiding the convergence issue altogether. In the following sections, we represent the SRR normal equations as a multirate system and present the basic mathematical tools for its analysis.

\begin{figure}
    \centering
    \includegraphics[width=8cm]{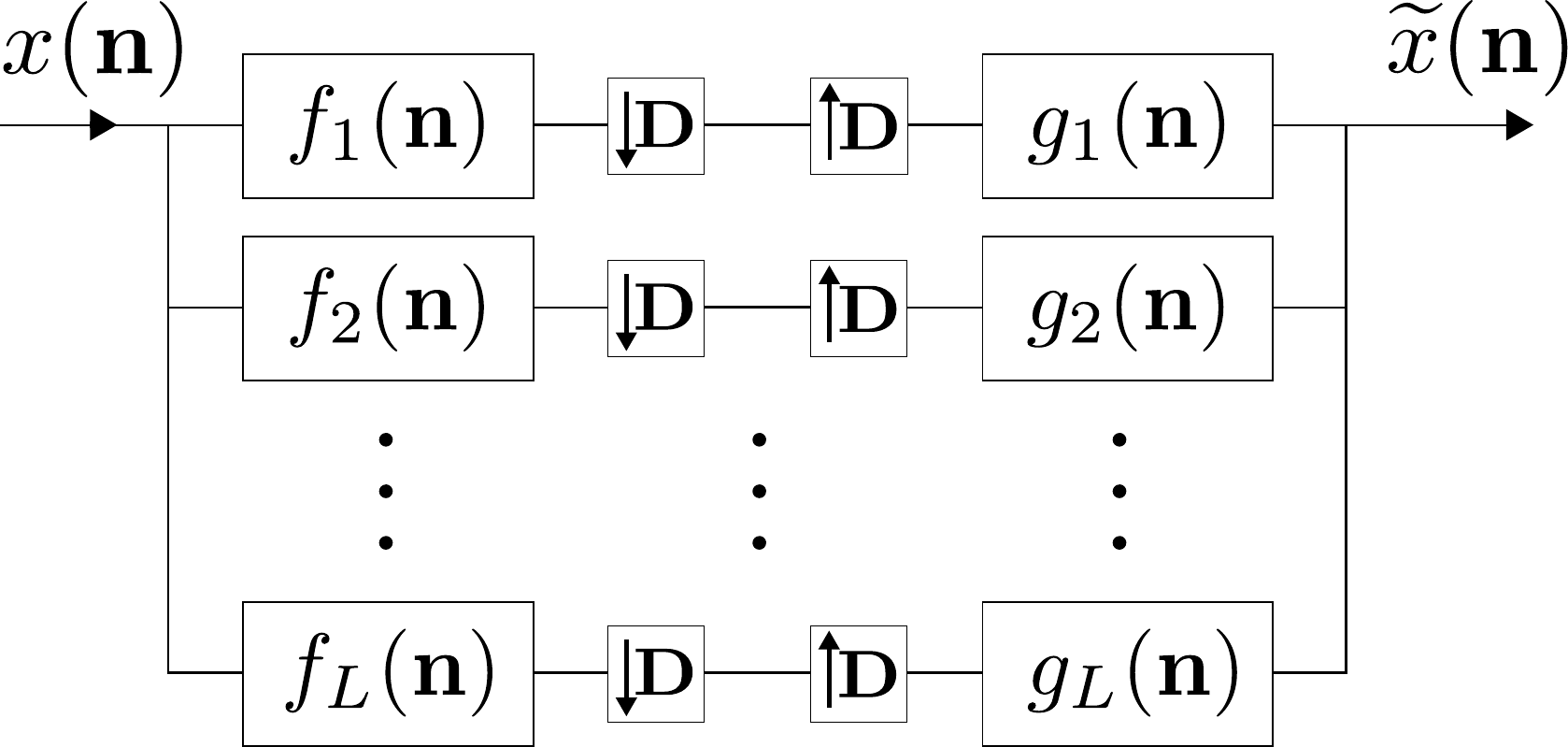}
    \caption{General form of a multidimensional multirate system.}
    \label{fig:standard_mrate_sys}
\end{figure}
\section{A Multirate Representation of Video SRR}
\label{sec:multirate_srr_cf}

Several image deblurring techniques are able to achieve a small and scalable computational complexity by modeling the image blurring process as a linear, shift invariant operation in the spatial domain\footnote{Spatially variable blur can also be considered, although the deblurring problem becomes significantly more complex~\cite{wang2014reviewDeblurring}.}~\cite{kruse2017limitsFFTdeblurringCNN}.
This characteristic allows the inverse problem to be solved efficiently by using the Fast Fourier Transform (FFT), which then becomes the main computational bottleneck of these methods~\cite{kruse2017limitsFFTdeblurringCNN}.
However, the presence of decimation operators in~\eqref{eq:sys_eq_trlms} precludes the direct use of FFT-based techniques to develop low-complexity solutions to the video SRR problem since the shift invariance property is lost.
Nevertheless, although no longer shift invariant, the presence of decimation can be handled using the theory of multirate systems, which provides us with tools analogous to those used for analyzing LSI systems to study systems composed of linear filter operations performed at multiple sampling rates.

In a general multirate system as depicted in Figure~\ref{fig:standard_mrate_sys}, an input signal is first processed by a set of $L$ linear filters $f_1,\ldots,f_L$, which are called analysis filters. Afterwards, the outputs of those filters are then downsampled and can undergo some form of processing depending on the application. Finally, the processed signals are then upsampled and filtered by another set of filters $g_1,\ldots,g_L$, which are called synthesis filters, and added together to form the output of the filterbank.
Multirate systems theory allows one to analyze and devise filterbanks with different structures for many applications, and can specify under which conditions the input signal in the filterbank can be perfectly reconstructed from its output.

By looking at the SRR normal equations~\eqref{eq:sys_eq_trlms} from the perspective of a multirate system/filterbank, we can make use of the underlying theory to provide an efficient solution to this problem without need to resort to more costly or inefficient iterative methods.
%
Due to the direct relationship between matrices $\mD$, $\mH$ and $\mS$ and spatial decimation and convolutions, the system matrix in the left hand side of equation~\eqref{eq:sys_eq_trlms} is equivalent to a multirate system, and can be easily represented in the form depicted in Figure~\ref{fig:standard_mrate_sys} by decomposing the spatial responses as

\begin{align} \label{eq:popyp_analysis_filts_SRR}
f_{\ell}(\vn) {}={}
\left\{
\begin{array}{ll}
    h(\vn) + s(\vn), & \ell=1 \\
    s(\vn) * \delta(\vn-\vk_{\ell}^{\text{SRR}}), & \ell=2,\ldots,d^2
\end{array}\right.
\end{align}
\begin{align} \label{eq:popyp_synth_filts_SRR}
g_{\ell}(\vn) {}={}
\left\{
\begin{array}{ll}
    h(-\vn) + s(-\vn), & \ell=1 \\
    s(-\vn) * \delta(\vn+\vk_{\ell}^{\text{SRR}}), & \ell=2,\ldots,d^2
\end{array}\right.
\end{align}
for $\ell=1,\ldots,L$, where $d^2=M^2/N^2$ is the decimation rate, $\delta(\vn)$ is the bidimensional Kronecker delta function (\textit{i.e.}, $\delta(\vn)=1$ if $\vn=\boldsymbol{0}$ and 0 otherwise) and $\vk_{\ell}^{\text{SRR}}\in\mathbb{R}^2$ are vectors are drawn from the set
\begin{align}
    \vk_{\ell}^{\text{SRR}} \in \bigg\{
    \left[\begin{array}{c} i \\ j 
    \end{array}\right], \,
    i,j=0,\ldots, d\bigg\}
    , \, \ell=1,\ldots, d^2.
\end{align}

In the following, we present the theoretical tools required to analyze multirate systems, which are based on the so-called polyphase transform. Afterwards, we compute an efficient solution to the SRR problem using this formulation, which is given in the form of a filterbank which approximately computes the inverse of the matrix in the left hand side of~\eqref{eq:sys_eq_trlms}. This methodology will be of great importance to guarantee the efficiency of the proposed edge-preserving SRR algorithm in Section~\ref{sec:edge_preserv}.





\section{Multidimensional Multirate Systems}
\label{sec:multirate_theory}

The characteristic that precludes the classical LSI tools to be applied to multirate systems consists on the lack of shift invariance due to the decimation operators.
The polyphase representation is a convenient way to deal with this issue by converting multirate systems into multi-input LSI system~\cite{kovacevic1992nonseparableMultidimFbank}.
In order to do so, we shall first introduce some necessary notation and definitions.

Let us first represent a generic HR image $\mX\in\mathbb{R}^{M\times M}$ as a bidimensional discrete signal $x(\vn):\mathbb{Z}^2\to\mathbb{R}$, which maps the discrete spatial positions to the pixel values (zero padding or other boundary conditions can be used to define the values of the pixels that are not contained in $\mX$). This allows the downsampling operation of a bidimensional signal (in the spatial domain) to be described using an integer matrix~$\mM\in\mathbb{Z}^{2\times2}$ as $x(\vn)\mapsto x(\mM\vn)$, where $x(\mM\vn)$ is the downsampled signal. Matrix $\mM$ determines which samples of $x(\vn)$ (for $\vn\in\mathbb{Z}^2$) are present in the downsampled signal $x(\mM\vn)$, and the constant $|\det(\mM)|=d^2$ is the downsampling factor at each channel~\cite{vaidyanathan1993multirateBook}\footnote{The decimation matrix for the SRR case presented in the previous section is given by $\mM=\left[\scalemath{0.7}{\begin{array}{cc}d&0\\0&d\end{array}}\right]$.}.

Using the downsampling matrix~$\mM$ and a set of vectors 
$\mathcal{N}(\mM)=\{\mM\vxi\in\mathbb{Z}^2\,:\,\vxi\in[0,1)^2\}=\{\vk_1,\ldots,\vk_{d^2}\}$, called the \emph{fundamental parallelepiped} of~$\mM$, the discrete spatial domain~$\mathbb{Z}^2$ can be decomposed into a set of $d^2$ \emph{polyphase components} of the form $\{\mM\vn + \vk_i : \vn\in\mathbb{Z}^2\}$, for $i=1,\ldots,d^2$~\cite{kovacevic1992nonseparableMultidimFbank,vaidyanathan1993multirateBook}, which allows us to write $\mathbb{Z}^2=\bigcup_{i=1}^{d^2}\{\mM\vn + \vk_i : \vn\in\mathbb{Z}^2\}$.



Using the polyphase decomposition, the input signal $x(\vn)$, the analysis filters $f_{\ell}(\vn)$, and the synthesis filters $g_{\ell}(\vn)$ in Figure~\ref{fig:standard_mrate_sys} can be decomposed in the Z-domain as
\begin{align} \label{eq:polyp_dec_i}
\begin{split}
    X_{\vk_i}(z) & {}={} \sum_{\vn\in\mathbb{Z}^2} x(\mM\vn-\vk_i)\,z^{-\vn}
    \\
    F_{\ell,\vk_i}(z) & {}={} \sum_{\vn\in\mathbb{Z}^2} f_{\ell}(\mM\vn + \vk_i) \, z^{-\vn}
    , \,\, \ell=1,\ldots,L
    \\
    G_{\ell,\vk_i}(z) & {}={} \sum_{\vn\in\mathbb{Z}^2} g_{\ell}(\mM\vn - \vk_i) \, z^{-\vn}
    , \,\, \ell=1,\ldots,L
\end{split}
\end{align}
where $X_{\vk_i}(z)$, $F_{\ell,\vk_i}(z)$ and $G_{\ell,\vk_i}(z)$ are the $i^{\rm th}$ polyphase components of the input signal, analysis and synthesis filterbanks, respectively.

Note that the signals and filters in~\eqref{eq:polyp_dec_i}  can be recovered from their polyphase components as
\begin{align} \label{eq:multirate_recover_filts}
\begin{split}
    X(z) & {}={} 
    \sum_{\vk\in\mathcal{N}(\mM)}
    z^{\vk} X_{\vk}(z^{\mM})
    \\
    F_{\ell}(z) & {}={} \sum_{\vk\in\mathcal{N}(\mM)} z^{-\vk} F_{\ell,\vk}(z^{\mM})
    , \,\, \ell=1,\ldots,L
    \\
    G_{\ell}(z) & {}={} \sum_{\vk\in\mathcal{N}(\mM)} z^{\vk} G_{\ell,\vk}(z^{\mM})
    , \,\, \ell=1,\ldots,L
\end{split}
\end{align}
where $X(z)$, $F_{\ell}(z)$ and $G_{\ell}(z)$ denote the Z-transform of $x(\vn)$, $f_{\ell}(\vn)$ and $g_{\ell}(\vn)$, and $z^{\mM}=(z_1^{\mM_{11}}z_2^{\mM_{21}},z_1^{\mM_{12}}z_2^{\mM_{22}})$~\cite{kovacevic1992nonseparableMultidimFbank}.

We denote by $\vx_p(z)=[X_{\vk_1}(z),\ldots,X_{\vk_{d^2}}(z)]$, $\vf_{p,\ell}(z) = [F_{\ell,\vk_1}(z),\ldots,F_{\ell,\vk_{d^2}}(z)]$ and $\vg_{p,\ell}(z) = [G_{\ell,\vk_1}(z),\ldots,G_{\ell,\vk_{d^2}}(z)]$ the row vectors containing all the polyphase components of the input signal, analysis and synthesis filters, respectively.

The polyphase components of the analysis and synthesis filters can be represented in matrix form by combining the components of the~$L$ filters as~\cite{kovacevic1992nonseparableMultidimFbank}
\begin{align}
    \mF_p(z) = \left[\begin{array}{cc} \vf_{p,1}(z) \\ \vdots \\ \vf_{p,L}(z)  \end{array}\right]
    ,\quad 
    \mG_p(z) = \left[\begin{array}{cc} \vg_{p,1}(z) \\ \vdots \\ \vg_{p,L}(z)  \end{array}\right]^\top
\end{align}
Note that each component of the polyphase matrix of the synthesis filters $\mG_p(z)$ is defined in a reverse order to that of the analysis filters $\mF_p(z)$.

Considering the representation of the signals and filters in the polyphase domain, the output of the filterbank is given by~\cite{kovacevic1992nonseparableMultidimFbank}:
\begin{align}
    \widetilde{X}(z) = \vp(z)\mG_p(z^{\mM})\mF_p(z^{\mM}) \vx_p(z^{\mM})
    \,\text{.}
\end{align}
where $\widetilde{X}(z)$ is the Z-transform of the output of the filterbank $\widetilde{x}(\vn)$, and $\vp(z)=[z^{\vk_1},\ldots,z^{\vk_{d^2}}]$~\cite{kovacevic1992nonseparableMultidimFbank}.

The polyphase transfer matrix describes the input-output relationship of a multirate system in the polyphase domain. It is obtained as the composition of the analysis and synthesis polyphase matrices, and is given by $\mT(z)=\mG_p(z)\mF_p(z)$.
Note that for a case such as the video SRR problem presented in Section~\ref{sec:TSR_LMS}, the polyphase matrix $\mT(z)$ corresponds to the polyphase representation related to the matrix at the left hand side of~\eqref{eq:sys_eq_trlms}, and is obtained by considering the multirate system in equations~\eqref{eq:popyp_analysis_filts_SRR} and~\eqref{eq:popyp_synth_filts_SRR}.


\subsection{An Inverse Polyphase Filter}
\label{sec:inv_filterbank_thr}

An important problem consists of designing a polyphase system $\mU(z)$ which can invert the response of a given polyphase system $\mT(z)$.
This is equivalent to the problem of finding a Laurent polynomial matrix $\mU(z)$ such that
\begin{align} \label{eq:fbank_inv_condition}
    \mU(z)\mT(z)=\mI
\end{align}
where each entry of $\mU(z)$ is required to be a Laurent polynomial, guaranteeing that the inverse filters are of FIR. This way, any input signal $\vz$ in the polyphase domain is perfectly recovered at the output since $\mU(z)\mT(z)\vz=\vz$.

However, designing the inverse of multidimensional multirate systems is not trivial.
The existence of solutions $\mU(z)$ to equations of the form of~\eqref{eq:fbank_inv_condition} has already been studied in the literature. Specifically, it has been found that a Laurent polynomial left inverse can (generically) only be obtained when $L-d^2\geq2$, where $d^2$ is the number of polyphase components of the transform~\cite{law2009genericInvertibilityFbank}.
%
%
This condition cannot be satisfied when $\mT(z)$ corresponds to the video SRR system matrix in~\eqref{eq:sys_eq_trlms} for any integer decimation factor $d>1$. Therefore, an exact left inverse will almost surely have rational entries, leading to IIR filters.

Nevertheless, we can still design a finite impulse response (FIR) polyphase filterbank that approximates the inverse of $\mT(z)$ good enough for practical purposes (i.e. $\mU(z)\mT(z)\approx\mI$). These are sometimes called near-perfect reconstruction filterbanks, and their design have been extensively studied in the literature, where different additional criteria such as e.g. maximum passband ripple or transition band width was often imposed to the estimated filters~\cite{johnston1980QMFfilterbanks,bregovic2003PHDoptimalFilterbanks,bregovic2003OptimizationFilterbanks,pirani1984analyticalQMFfilterbanks}.
Note, however, that these additional constraints often result in complex or non-convex objective functions that are difficult to optimize~\cite{kumar2013filterbankLevenbergMarquardt,ho2010filterbankGlobalNonconvex}.


In this work, we consider a simple objective of designing a Laurent polynomial $\mU(z)$ that best approximates the inverse of $\mT(z)$ in the squared norm sense, i.e.,
\begin{align} \label{eq:filterbank_inv_2}
    & \min_{\mU(z)} \,\, \|\mU(z)\mT(z) - \mI\|_{\mathcal{L}}^2 
    \\[-1.5ex] & \nonumber 
    \qquad\quad {}={} \min_{\mU(z)} \,\, \sum_{i,j=1}^{d^2} \int_{\|z\|=1} 
    \hspace{-1ex} \big|[\mU(z)\mT(z) - \mI]_{(i,j)}\big|^2 dz
\end{align}
where $[\,\cdot\,]_{(i,j)}$ denotes the $(i,j)^{\rm th}$ position of a matrix. Note that the optimization problem in~\eqref{eq:filterbank_inv_2} is convex, and can also be expressed equivalently in the spatial domain as
\begin{align} \label{eq:filterbank_inv_3}
    \min_{U_{i,j}(\vn)}  \,\,  & \sum_{i=1}^{d^2} \bigg\| \delta(\vn) 
    - \sum_{m=1}^{d^2} U_{i,m}(\vn) * T_{m,i}(\vn) \bigg\|_F^2
    \nonumber \\ & 
    +
    \sum_{i,j=1,\,i\neq j}^{d^2}
    \bigg\| \sum_{m=1}^{d^2} U_{i,m}(\vn) * T_{m,j}(\vn) \bigg\|_F^2
\end{align}
where $*$ is the bidimensional convolution operator, and $T_{i,j}(\vn)$ and $U_{i,j}(\vn)$ are the inverse Z-transform of $\big[\mT_{i,j}(z)\big]_{(i,j)}$ and $\big[\mU_{i,j}(z)\big]_{(i,j)}$, respectively.

\subsection{Video SRR using an inverse polyphase filter}
\label{sec:invFb_srr}

The video SRR problem in~\eqref{eq:sys_eq_trlms} can be solved with good accuracy by using an approximate inverse filterbank computed using the approach described in Section~\ref{sec:inv_filterbank_thr}, with $\mT(z)$ corresponding to the polyphase representation of the matrix at the left hand side of~\eqref{eq:sys_eq_trlms}. We call this solution the \textit{Multirate Temporally Selective Regularized LMS} (MTSR-LMS) algorithm.

After solving \eqref{eq:filterbank_inv_2} or \eqref{eq:filterbank_inv_3}, the resulting polyphase transfer matrix $\mU(z)$ can be factored into the polyphase matrices of analysis and synthesis filterbanks as $\mF_p'(z)=\mI$ and $\mG_p'(z)=\mU(z)$. From these matrices, the set of $L=d^2$ multirate filters $f_1'(\vn),\ldots,f_{d^2}'(\vn)$ and $g_1'(\vn),\ldots,g_{d^2}'(\vn)$
(corresponding to those in Figure~\ref{fig:standard_mrate_sys}) can be obtained by using equation~\eqref{eq:multirate_recover_filts} and the inverse Z-transform.


After these filters are computed, the solution $\vxh(k)$ to the video SRR problem~\eqref{eq:sys_eq_trlms} is then obtained by applying the multirate filterbank to the signal $\mH^\top\mD^\top\vy(k)+\mS^\top\mS\mG(k)\vxh(k-1)$.

%

\section{Efficient Edge-Preserving Video SRR}
\label{sec:edge_preserv}

Although the multirate-based solution to problem~\eqref{eq:sys_eq_trlms} provides a very efficient video SRR method with considerable robustness to innovations, the underlying formulation of the algorithm in equation \eqref{eq:Lagrangian2} is based on a Thikhonov or L$_2$-norm-based spatial regularization, which tends to overly smooth image edges.
As discussed in Section~\ref{sec:TSR_LMS}, recent works in SRR, image restoration and denoising address this issue by using edge-preserving regularization strategies. Those include the classical~\cite{farsiu2004fastRobustSRR,liu2014bayesianVideoSRR}, reweighted~\cite{kohler2016robustSRRirsw} and Wavelet-based~\cite{figueiredo2003EM_waveletRestoration,figueiredo2007MM_waveletRestoration,afonso2010fastImgRecVariableSplitting,matakos2013acceleratedRestoration} versions of the Total Variation penalty, which aims to promote piecewise smooth images, and more costly methods that explore non-local spatial redundancy~\cite{buades2005reviewDenoisingNLM}.
%
%
%

In the following, we will consider an edge preserving spatial regularization by modifying the video SRR cost function in~\eqref{eq:Lagrangian2} to include a sparsity promoting penalty of the Wavelet-domain representation of the estimated image~\cite{figueiredo2003EM_waveletRestoration,figueiredo2007MM_waveletRestoration,afonso2010fastImgRecVariableSplitting,matakos2013acceleratedRestoration}.
%
This corresponds to a regularization term of the form $\mathcal{R}(\vxh(k))=\|\mW\vxh(k)\|_p$, where $\mW\in\mathbb{R}^{M\times M}$ is the discrete Wavelet transform matrix (which satisfies $\mW^\top\mW=\mI$) and~$p$ is either~$0$ or~$1$. 

In this case, the video SRR problem becomes:
\begin{align} \label{eq:edgePreserving1}
\begin{split}
    \min_{\vxh(k)} \,\,\, & \| \vy(k)-\mD\mH \vxh(k) \|^{2}_2
	+ \alpha \|\mW\vxh(k)\|_p
	\\ &
	+ \alpha_{\tiny{\text{T}}} \big\|\mS\big[\vxh(k) -\mG(k)\vxh(k-1)\big]\big\|^2_2.
\end{split}
\end{align}
%
We consider the Wavelet-based regularization strategy instead of other alternatives such as the Total Variation due to the orthogonality property of the Wavelet transform, which will be explored in order to devise an efficient solution to problem~\eqref{eq:edgePreserving1}.
%
Note that although~\eqref{eq:edgePreserving1} becomes a non-convex optimization problem when $p=0$, it has been observed that this choice of $p$ (\textit{e.g.} in hard threshonding algorithms) usually leads to a better preservation of image edges when compared to $p=1$~\cite{buades2005reviewDenoisingNLM}.

Although the cost function in~\eqref{eq:edgePreserving1} usually leads to reconstructed images with better perceptual quality, minimizing it proves to be much more difficult.
Since there are no closed form solutions to~\eqref{eq:edgePreserving1}, iterative optimization procedures must be employed, such as iterative shrinkage/thresholding~\cite{figueiredo2003EM_waveletRestoration,daubechies2004sparse_ISTA,beck2009sparse_FISTA,beck2009denoisingDeb_FISTA,matakos2013acceleratedRestoration} or variable splitting~\cite{ng2010TV_admm,boyd2011ADMM,afonso2010fastImgRecVariableSplitting,goldstein2009splitBregmanL1,afonso2011ADMM_imagingInvProb} methods.
However, these techniques also result in a very high computational complexity when compared to video SRR solutions available for quadratic regularization terms, jeopardizing the possibility of an efficient algorithm.
Therefore, alternative approaches are needed in order to provide solutions which are more suitable for real-time implementation.

In order to achieve this goal, we propose to formulate the video SRR problem~\eqref{eq:edgePreserving1} in the form of set theoretic estimation~\cite{combettes1993foundationsSetThrEstim}. This allows us to employ an alternating projection approach, which provides reasonable reconstruction performance at a very small number of iterations, allowing for real-time implementation. 

Set theoretic estimation formulates an estimation problem such that its solutions are contained within the intersection of several sets in the solution space, where each of those sets represents one piece of information about the problem~\cite{combettes1993foundationsSetThrEstim}.
More precisely, we consider sets $\Omega_1,\Omega_2,\ldots,\Omega_R$ with nonempty intersection $\Omega=\cap_{r=1}^R\Omega_r\neq\varnothing$. The solutions to the feasibility problem are any point $x\in\Omega$. Various strategies exist to compute $x\in\Omega$, a simple one being an alternating projection approach. It can be described in its most general form as follows. Given an initial solution $x_0$, it is updated as \cite{combettes1993foundationsSetThrEstim}
\begin{align} \label{eq:altProj0}
    x_{j+1}=\mathcal{P}_{\iota_j}(x_j), \,\,
    \iota_j\in\{1,\ldots,R\}, \,\,
    j\in\mathbb{Z}_+
\end{align}
where $\mathcal{P}_r(\cdot)$ is a projection operation onto set $\Omega_r$, and $\iota_j$ is a sequence which determines the sequence of sets considered in the projections.
If $\Omega_r$ are convex and closed sets in a Hilbert space, then the sequence $\{x_j\}$ converges to a point $x\in\Omega$ as long as the sequence of indices $\iota_j$ is cyclic and contains all elements of $\{1,\ldots,R\}$~\cite{bregman1967projectionsConvexSets}.

When the sets $\Omega_r$ are non-convex, however, algorithm~\eqref{eq:altProj0} is not always guaranteed to converge. Nevertheless, convergence has still been established under different conditions~\cite{lewis2009locConvergenceAltProjNonConvex,zhu2018convergenceAltProjNonConvex}, including particular cases of the sparse regression problem based on the $L_0$ semi-norm~\cite{lai2017convergenceAltProjSparse} which matches our spatial regularization for $p=0$ and for which alternating projection approaches have been successfully applied in practice~\cite{portilla2007alternatingProjectionL0}.


For the video SRR problem, we decompose the cost function in~\eqref{eq:edgePreserving1} into two sets in the solution space. The first one describing the solutions $\vxh(k)$ which agree with the present observation $\vy(k)$ and whose details are consistent with the registered previous estimate $\mG(k)\vxh(k-1)$. The second set then describes the solutions $\vxh(k)$ which have bounded variation or which are sparse in the Wavelet domain. 
More precisely, they are given by
\begin{align}
\begin{split}
    \Omega_1 & {}={} \big\{\vxh(k) \,:\, \|\vy(k)-\mD\mH\vxh(k)\|^2_2\leq\sigma, \\ 
    & \hspace{1.7cm} \big\|\mS\big[\vxh(k)-\mG(k)\vxh(k-1)\big]\big\|^2_2\leq\gamma \big\} \\
    \Omega_2 & {}={} \big\{\vxh(k) \,:\, \|\mW\vxh(k)\|_p\leq\tau \big\} \\
\end{split}
\end{align}
where parameters $\sigma$, $\gamma$ and $\tau$ control the diameter of the sets $\Omega_1$ and $\Omega_2$, and are indirectly related to the regularization parameters.

Given a solution at the $(j-1)^{\rm th}$ iteration $\vxh_{j-1}(k)$, the successive projections at iteration $j$ are given by
\begin{align} \label{eq:altProj1}
\begin{split}
    \vz & {}={} \mathop{\arg\min}_{\vx \in \Omega_1} \,\, \|\vx - \vxh_{j-1}(k)\|_2    \\
    \vxh_{j}(k) & {}={} \mathop{\arg\min}_{\vx \in \Omega_2} \,\, \|\vx - \vz\|_2.
\end{split}
\end{align}

In order to simplify these problems, we convert them into unconstrained optimization problems. These optimization problems are equivalent to
\begin{align} \label{eq:altProj2}
\begin{split}
    & \vz = \mathop{\arg\min}_{\vx}  \, \Big\{\|\vx - \vxh_{j-1}(k)\|^2_2 + \lambda_1 \big(\|\vy(k)-\mD\mH\vx\|^2_2 
    \\ & \hspace{2cm}
    + \alpha_{\tiny{\text{T}}} \big\|\mS\big[\vx-\mG(k)\vxh(k-1)\big]\big\|^2_2 \big)\Big\}
    \\
    & \vxh_{j}(k) = \mathop{\arg\min}_{\vx}  \|\vx - \vz\|^2_2 + \lambda_{\tau} \|\mW\vx\|_p
\end{split}
\end{align}
in the sense that there exist bijections between parameters $(\sigma,\gamma)$ and $(\lambda_1,\alpha_{\tiny{\text{T}}})$, and between parameters $\tau$ and $\lambda_\tau$ such that the solutions to the problems in~\eqref{eq:altProj1} and~\eqref{eq:altProj2} are the same~\cite{vaiter2015regularizationInverseProblems}. These bijections, however, are not explicit and depend on variables such as $\vxh_{j-1}(k)$, so these problems are usually treated individually from a computational point of view~\cite{vaiter2015regularizationInverseProblems}.
Since the parameters $(\sigma,\gamma,\tau)$ are usually not known in advance, we instead focus on the problems in~\eqref{eq:altProj2} and directly select parameters $(\lambda_1,\alpha_{\tiny{\text{T}}},\lambda_\tau)$ empirically. 

The solutions to the optimization problems in~\eqref{eq:altProj2} can be computed very efficiently, and are given by
\begin{subequations}
\begin{align} 
\begin{split} \label{eq:alg_fbWav_projSol_1}
    & \vz {}={} \big(\lambda_1\mH^\top\mD^\top\mD\mH 
    + \lambda_1\alpha_{\tiny{\text{T}}} \mS^\top\mS + \mI\big)^{-1} \big[
    \vxh_{j-1}(k) 
    \\ & \qquad + \lambda_1\mH^\top\mD^\top\vy(k) 
    + \lambda_1\alpha_{\tiny{\text{T}}}\mS^\top\mS\mG(k)\vxh(k-1) \big]
\end{split}
    \\ \label{eq:alg_fbWav_projSol_2} 
    & \vxh_{j}(k) {}={} \mW^\top \operatorname{thr}_p(\mW\vz,\lambda_\tau)
\end{align}
\end{subequations}
where $\operatorname{thr}_p(\mW\vz,\lambda_\tau)$ is a  soft thresholding operation for $p=1$~\cite{beck2009sparse_FISTA}, given by
\[\operatorname{thr}_1(\mW\vz,\lambda_{\tau}) {}={}
\max\big(|\mW\vz| - \lambda_\tau,\mathbf{0}\big)\,\operatorname{sign}(\mW\vz) \,,\]
and a hard thresholding operator for $p=0$~\cite{buades2005reviewDenoisingNLM}, given by
\[[\operatorname{thr}_0(\mW\vz,\lambda_\tau)]_i= 
\left\{\begin{array}{cc}
    [\mW\vz]_i & [\mW\vz]_i \geq \lambda_\tau \\
    0 & [\mW\vz]_i < \lambda_\tau
\end{array}\right.
\]
for $i=1,\ldots,M$, where $[\,\cdot\,]_i$ denotes the $i^{\rm th}$ position of a vector.

Note that the matrix inverse in equation~\eqref{eq:alg_fbWav_projSol_1} can be computed very efficiently using the filterbank strategy proposed in the previous section.
%
Note also that the use of orthogonal transforms for $\mW$ (as in the case of the Wavelet transform) usually results in ringing artifacts in the reconstructed images. In order to overcome this issue, we can use the cycle spinning technique proposed by Coifman and Donoho~\cite{coifman1995denoisingCycleSpinning}. This technique adds redundancy to $\mW$ by making it shift invariant while also being strongly related to the TV regularization for image denoising~\cite{kamilov2012waveletCycleSpinningGeneralizeTV}, and is now largely used to improve the quality of these kind of methods~\cite{vonesch2008fastWaveletCycleSpinning,figueiredo2003EM_waveletRestoration,guerquin2011waveletCycleSpinningMRI}.

Although the solutions to problems in~\eqref{eq:altProj2} can be performed efficiently, alternating projection approaches are well known to suffer from slow convergence and a high computational cost~\cite{Park03}.
%
In order to overcome this problem, we propose to adjust the sizes of the sets during the iterations to accelerate convergence and maintain robustness to innovations. 
We do so by controlling $\lambda_1\equiv\lambda_1(j)$, which turns out to be the determinant parameter in the algorithm's convergence speed. The remaining parameters are maintained constant.
We select a large value for $\lambda_1(j)$ in the first iterations to reduce the effect of initialization, making the set $\Omega_1$ very small or even pointwise. This value can then be reduced in the subsequent iterations, yielding the rule $\lambda_1(j+1)\leq\lambda_1(j)$.
As long as $\lambda_1(j)$ converges to a fixed value after a finite number of iterations, the convergence properties of algorithm~\eqref{eq:altProj0} are maintained.

This strategy, however, must be adopted with care if we want to employ the inverse filterbank strategy of Section~\ref{sec:invFb_srr} to solve~\eqref{eq:alg_fbWav_projSol_1}, since each parameter $\lambda_1(j)$ leads to a different matrix inverse, and therefore requires a different filterbank to be solved. These filterbanks must be precomputed in order to have an efficient algorithm.
Thus, we only consider a set $\Lambda$ with small cardinality of possible values, satisfying $\lambda_1(j)\in\Lambda$ for $j=1,\ldots,J$. 
We use a simple parameter update rule in our experiments, with $\Lambda=\{\infty,1\}$ and we consider $\lambda_1(1)=\infty$, and $\lambda_1(j)=1$, $\forall j\geq2$.

This solution, which call the \textit{Wavelet-based Multirate Temporally Selective RLMS} (WMTSR-LMS) algorithm, is detailed in Algorithm~\ref{alg:algorithm_2}.

\begin{algorithm} [bth]
\small
\SetKwInOut{Input}{Input}
\SetKwInOut{Output}{Output}
\caption{WMTSR-LMS Algorithm~\label{alg:algorithm_2}}
\Input{Parameters $p\in\{0,1\}$, $\lambda_1(1),\ldots,\lambda_1(J)\in\Lambda$, $\lambda_{\tau}$, and $\alpha_{\tiny{\text{T}}}$ and matrices $\mH$, $\mD$, $\mS$.}
\Output{The super-resolved image sequence $\vxh(k)$, $k\in\mathbb{Z}_+$.}
Set $\lambda_1(1)=\infty$ \;
\For{$k\in\mathbb{Z}_+$}{
Estimate $\mG(k)$ from $\vy(k)$ and $\vy(k-1)$ \;
Initialize $\vxh_0(k)=\mG(k)\vxh(k-1)$ \;
\For{$j=1,\ldots,J$}{
Compute $\vz$ using equation~\eqref{eq:alg_fbWav_projSol_1} and $\lambda_1\equiv\lambda_1(j)$ \;
Compute $\vxh_{j}(k)$ using equation~\eqref{eq:alg_fbWav_projSol_2}  \;
}
Set $\vxh(k)=\vxh_J(k)$ \;
}
\end{algorithm}

\subsection{Computational Complexity}

In the MTSR-LMS algorithm, which considers a Thikonov spatial regularization, the main computational operations consist of computing the signal $\mH^\top\mD^\top\vy(k)+\mS^\top\mS\mG(k)\vxh(k-1)$ and passing it through a polyphase filter, which amounts to one convolution for each polyphase component of the input signal.
Thus, the total amount of operations is given by:
\[2|h(\vn)|M + 2|s(\vn)|M + 2 d^2\mathcal{O}(M\log_2(M))\]
where $|h(\vn)|$ and $|s(\vn)|$ denote the cardinality of the filters $h(\vn)$ and $s(\vn)$.

The WMTSR-LMS algorithm using the edge preserving regularization consists basically of evaluating the two expressions in equations~\eqref{eq:alg_fbWav_projSol_1} and~\eqref{eq:alg_fbWav_projSol_2} at each iteration $j$ of the alternating projection procedure.
The expression~\eqref{eq:alg_fbWav_projSol_1} is evaluated by applying a polyphase filterbank to an input signal, which has complexity of
\[2|h(\vn)|M + 2|s(\vn)|M + 2 d^2\mathcal{O}(M\log_2(M))\]
operations (\textit{i.e.} the same complexity of the MTSR-LMS algorithm).
Equation~\eqref{eq:alg_fbWav_projSol_2} involves the computation of a forward and inverse Wavelet transform, in addition to a thresholding operation with complexity $\mathcal{O}(M)$.

By considering separable Wavelet filters, the computational complexity of the translation invariant Wavelet transform~\cite{nason1995stationaryWavelet} with $Q$ decomposition levels consists of $4Q$ convolutions with rank-1 filters. Assuming the image to be square for simplicity, this amounts to a complexity of 
\[8 Q  \mathcal{O}(M\log_2(\sqrt{M})\]
operations.
By considering the overall cost for all iterations $j=1,\ldots,J$, the total number of operations for the WMTSR-LMS algorithm is given by
\[J\Big(2M(|h(\vn)| + |s(\vn)|) + (2 d^2 + 4 Q) \mathcal{O}(M\log_2(M)) + \mathcal{O}(M)\Big).\]


\section{Experimental Results}
\label{sec:results}

In this section, we present two examples to illustrate the performance of the proposed algorithms in terms of reconstruction quality and computational cost. 
The first example consists of a Monte Carlo (MC) simulation with synthetically generated video sequences, which aims to evaluate the proposed methods in terms of both robustness to innovations and reconstruction quality in a controlled environment.
Afterwards, the second example evaluates the performance of the algorithms when super-resolving real video sequences containing complex motion patterns and innovation outliers. For this example, we compare the proposed methods with state-of-the-art SRR algorithms, namely, a Bayesian method~\cite{liu2014bayesianVideoSRR} and a Convolutional Neural Network (CNN)~\cite{tao2017detailRevealingSRRneuralNets}.

In both examples, we also compare the proposed algorithms to a bicubic interpolation and to the LTSR-LMS SRR method~\cite{borsoi2017newSRRrobustnessInn,borsoi2019SRR_adap_TIP}, which uses a gradient descent approach to optimize the cost function in~\eqref{eq:Lagrangian2}.
For all simulations, the observed images were aligned using the \textit{Horn \& Schunck} registration algorithm~\cite{Horn1981determiningOF,Sun10}\footnote{The parameters were set as: \texttt{lambda=1$\times$10$^3$}, \texttt{pyramid\_levels=4}, \texttt{pyramid\_spacing=2}.}.
For the WMTSR-LMS algorithm, we considered $p=0$ (\textit{i.e.} L$_0$ (semi)-norm regularization) and used a Daubechies Wavelet with~5~vanishing moments,~4~decomposition levels and cycle spinning~\cite{coifman1995denoisingCycleSpinning}, and set the parameter $\lambda_{\tau}=10$.
To keep the computational complexity low, a single iteration per time instant (\textit{i.e.} $J=1$) was employed for both the MTSR-LMS and for the WMTSR-LMS, with the remaining parameter set as $\lambda_1(1)=\infty$ as discussed in the previous section.
All algorithms are initialized with a bicubic interpolation of the first LR image.


\begin{figure}
\centering
\includegraphics[width=8cm]{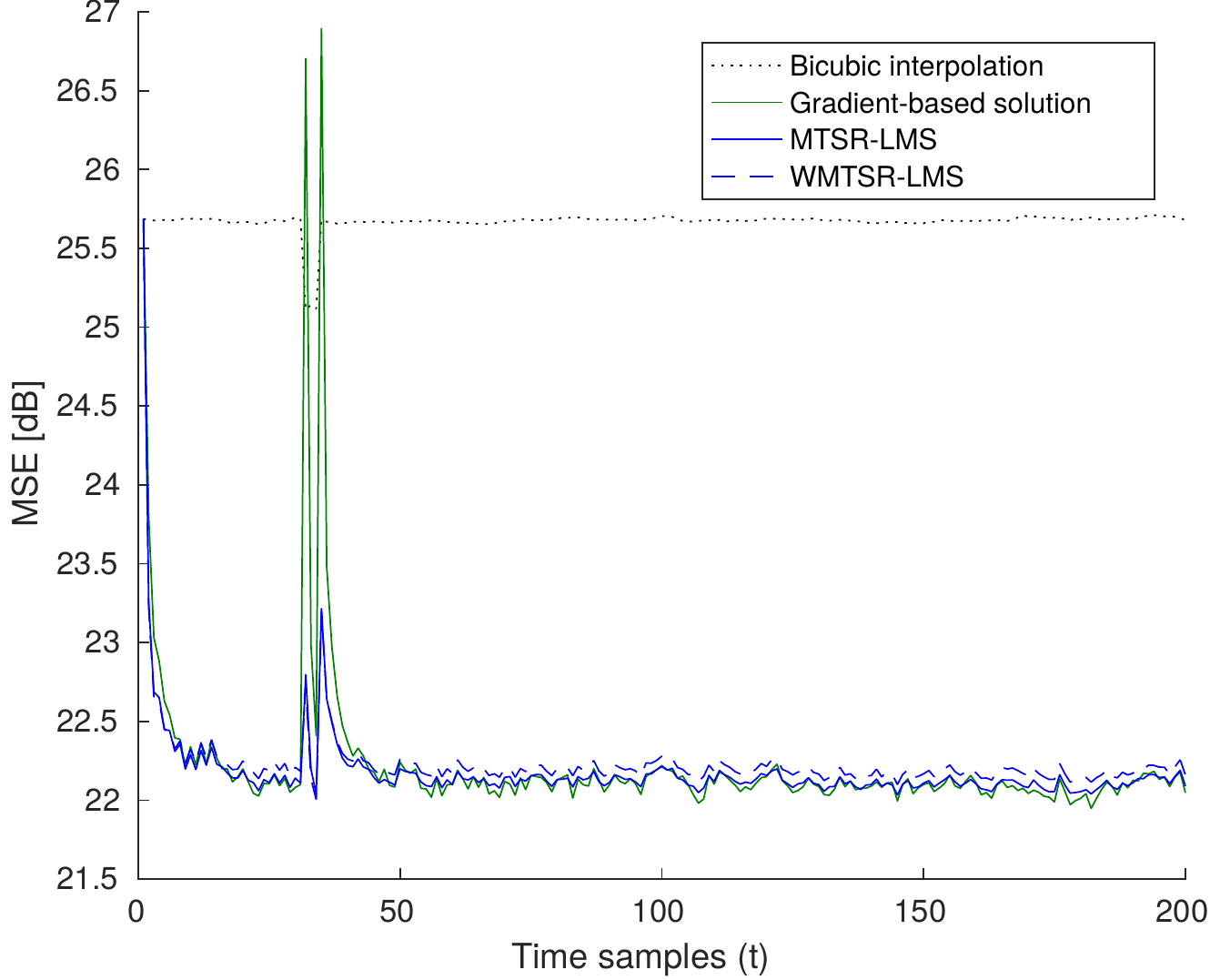}
\vspace{-0.2cm}
\caption{Average MSE per pixel for all algorithms.}
\label{fig:MSEknown}
\end{figure}

\begin{table}[th]
\footnotesize
\caption{Parameter values used in the simulations}
\vspace{-0.2cm}
\centering
\renewcommand{\arraystretch}{1.2}
\begin{tabular}{lccccc}
\hline
& $\mu$ & $\alpha$ & $\alpha_{\text{T}}$ & $J$ & $\lambda_{\tau}$ \\
\hline
Gradient Method~(LTSR-LMS~\cite{borsoi2019SRR_adap_TIP}) & $3.4$  & $10^{-4}$ &  $0.017$  & 2  & -- \\
\textbf{MTSR-LMS}     & --     & $0.005$   &  $0.015$  & -- & -- \\
\textbf{WMTSR-LMS}    & --     & --        &  $0.015$  & 1  & 10 \\
\hline
\end{tabular}
\label{tab:parametersOutlier}
\end{table}

\subsection{Example 1} \label{sec:examples_synth_i}

In this example, a Monte Carlo simulation was performed using synthetic video sequences. The HR video sequences were created by translating an $256\times256$ window with random, unitary displacements over distinct static images of natural scenes such as \textit{Lena}, \textit{Barbara} and others.
In order to emulate the behavior of an innovation (\textit{flying bird}) outlier in the HR sequences, a suddenly appearing object (independent of the background) was introduced in the video sequences, consisting of an $128\times128$ black square appearing in the middle of the~32nd frame and disappearing in the~35th frame of every sequence. This will allow us to assess the robustness of the algorithms.

The resulting HR sequences were then blurred with a uniform unitary gain $3\times3$ mask and decimated by a factor of~$2$, resulting in LR images of dimension $N=128$. Finally, white Gaussian noise with variance $\sigma^2=10$ was added to the decimated images. 
The performances of the algorithms were then evaluated using the MSE averaged over~30 input video sequences.

The parameters for the proposed algorithm and for the gradient-based solution~\cite{borsoi2019SRR_adap_TIP}, shown in Table~\ref{tab:parametersOutlier}, were selected in order to achieve minimum overall MSE and SSIM between frames~30 and~40. The average MSE/SSIM in this interval was estimated by running an exhaustive search over a small, independent set of synthetically generated video sequences and averaging the error metrics.

\begin{figure*}[htb]
\begin{minipage}[b]{.19\textwidth}
  \centering
  \centerline{\includegraphics[width=0.98\linewidth]{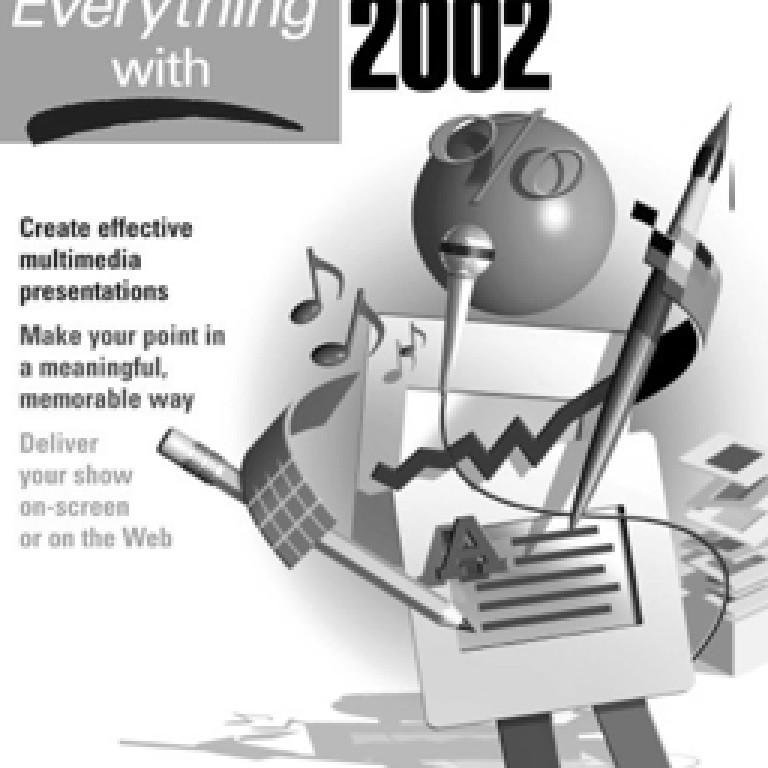}}
  \centerline{(a)}\medskip
\end{minipage}
\begin{minipage}[b]{.19\textwidth}
  \centering
  \centerline{\includegraphics[width=0.98\linewidth]{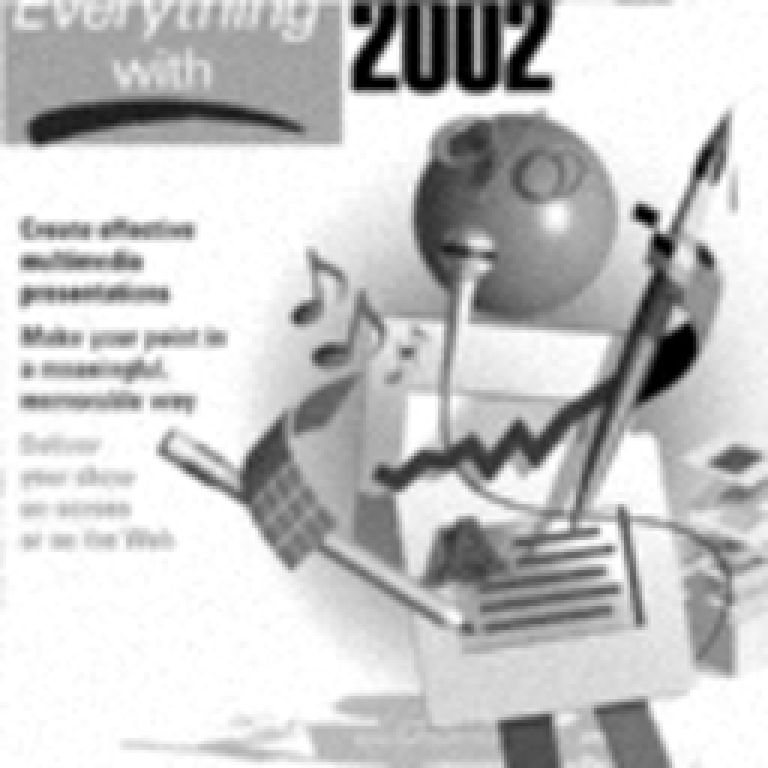}}
  \centerline{(b)}\medskip
\end{minipage}
\begin{minipage}[b]{.19\textwidth}
  \centering
  \centerline{\includegraphics[width=0.98\linewidth]{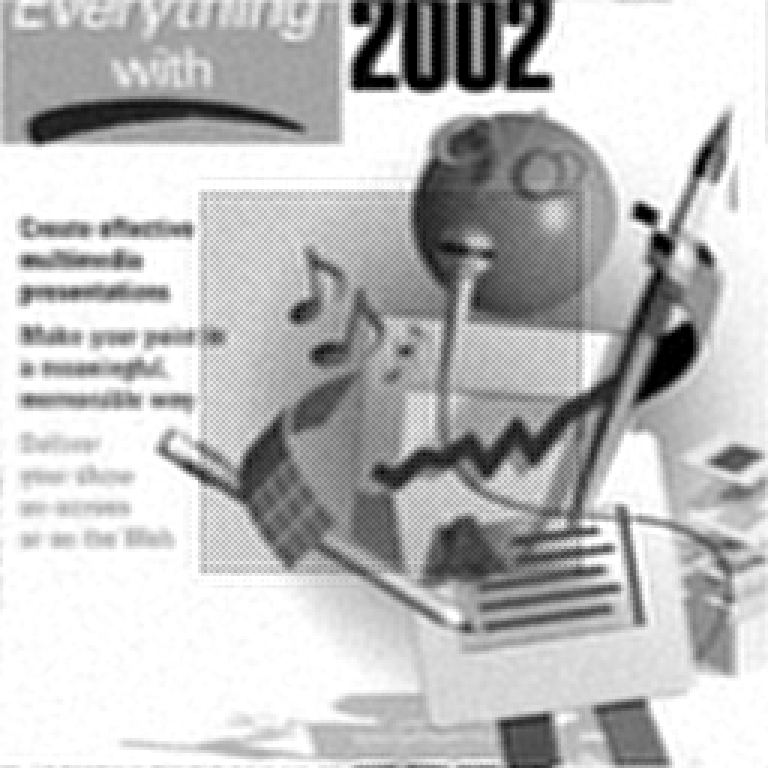}}
  \centerline{(c)}\medskip
\end{minipage}
%
\begin{minipage}[b]{.19\textwidth}
  \centering
  \centerline{\includegraphics[width=0.98\linewidth]{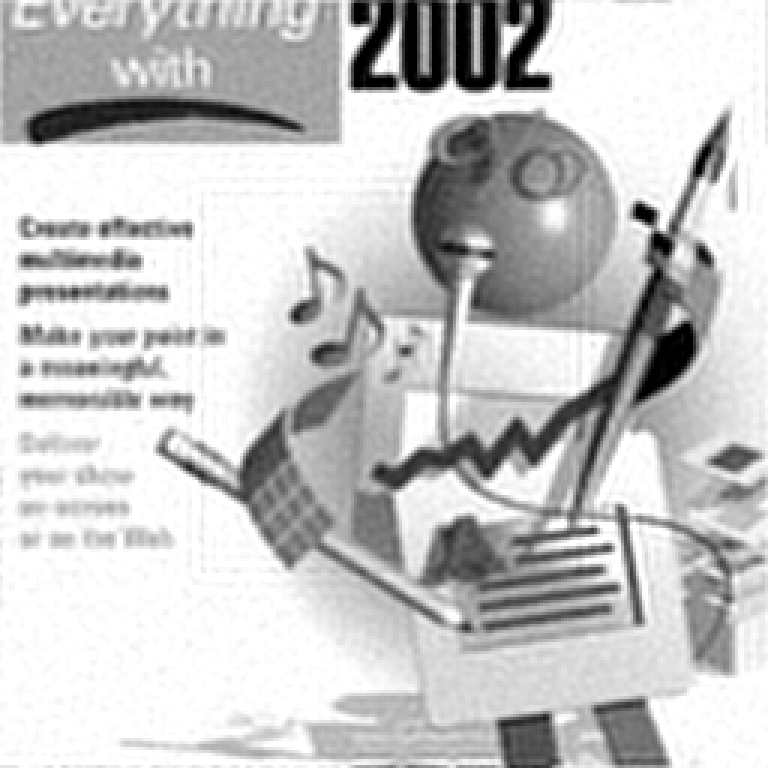}}
  \centerline{(d)}\medskip
\end{minipage}
\begin{minipage}[b]{.19\textwidth}
  \centering
  \centerline{\includegraphics[width=0.98\linewidth]{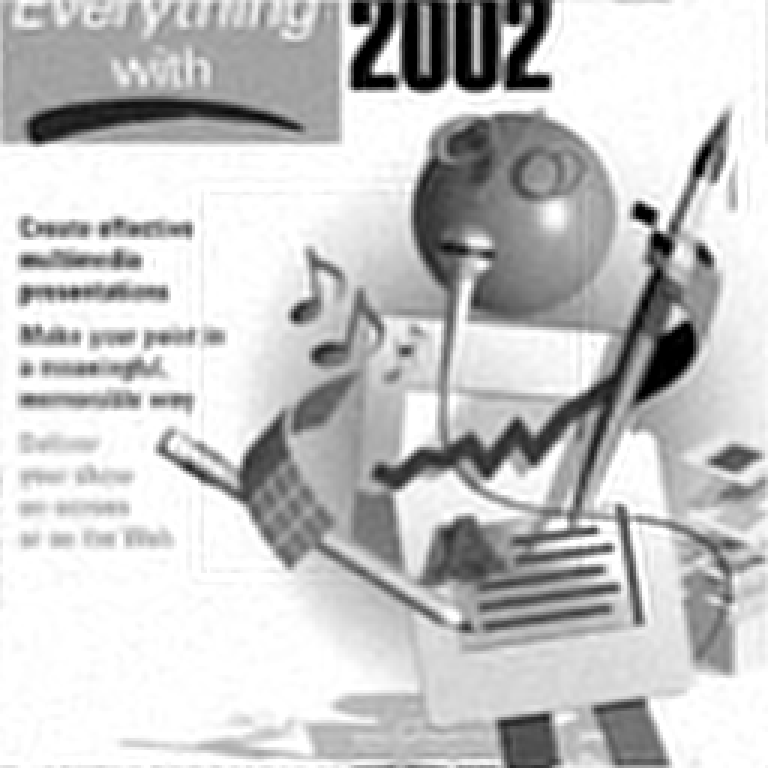}}
  \centerline{(e)}\medskip
\end{minipage}
\vspace{-0.4cm}
\caption{Sample of 35th frame of a reconstructed sequence. (a) Original image. (b) Bicubic interpolation (MSE=26.49dB, SSIM=0.800). (c) TSR-LMS (MSE=14.55dB, SSIM=0.732). (d) MTSR-LMS (MSE=13.87dB, SSIM=0.812). (e) WMTSR-LMS (MSE=13.86dB, SSIM=0.853).}
\label{fig:resBird}
\end{figure*}

The MSE evolution for all tested algorithms is presented in Figure~\ref{fig:MSEknown}.
It can be seen that the proposed methods are significantly more robust than the gradient-based solution (i.e. the LTSR-LMS algorithm), which shows significant spikes in the MSE between frames~32 and~35.
The reconstructed images for frame~35, shown in Figure~\ref{fig:resBird}, also support the quantitative results. In this frame, the black square (which is no longer present in the desired image sequence) is still clearly visible in the image reconstructed by the LTSR-LMS, as opposed to the proposed methods, which provide clear reconstruction results without significant artifacts.

\begin{figure*}[htb]
\begin{minipage}[b]{.19\textwidth}
  \centering
  \centerline{\includegraphics[width=0.98\linewidth]{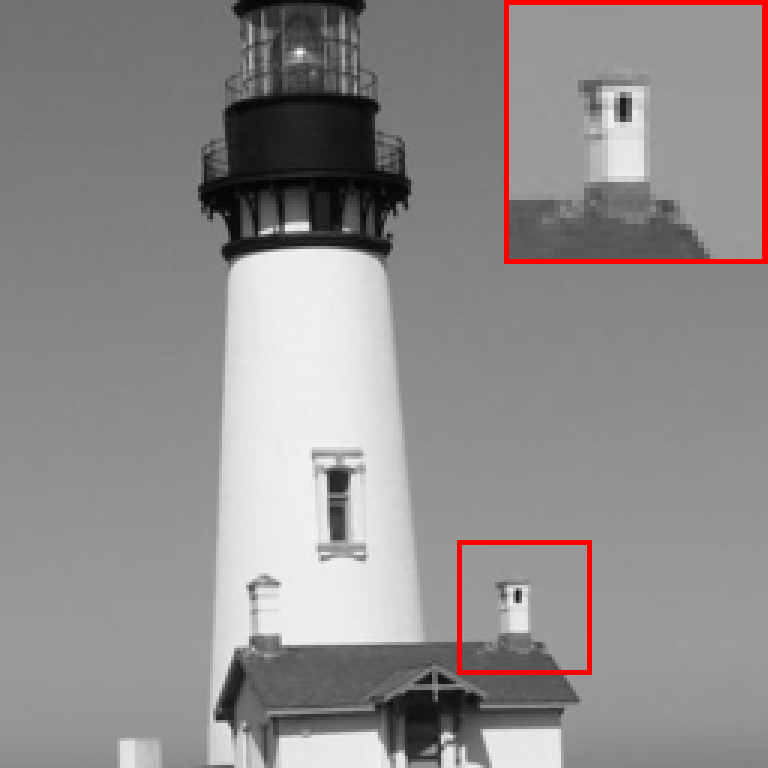}}
  \centerline{(a)}\medskip
\end{minipage}
\begin{minipage}[b]{.19\textwidth}
  \centering
  \centerline{\includegraphics[width=0.98\linewidth]{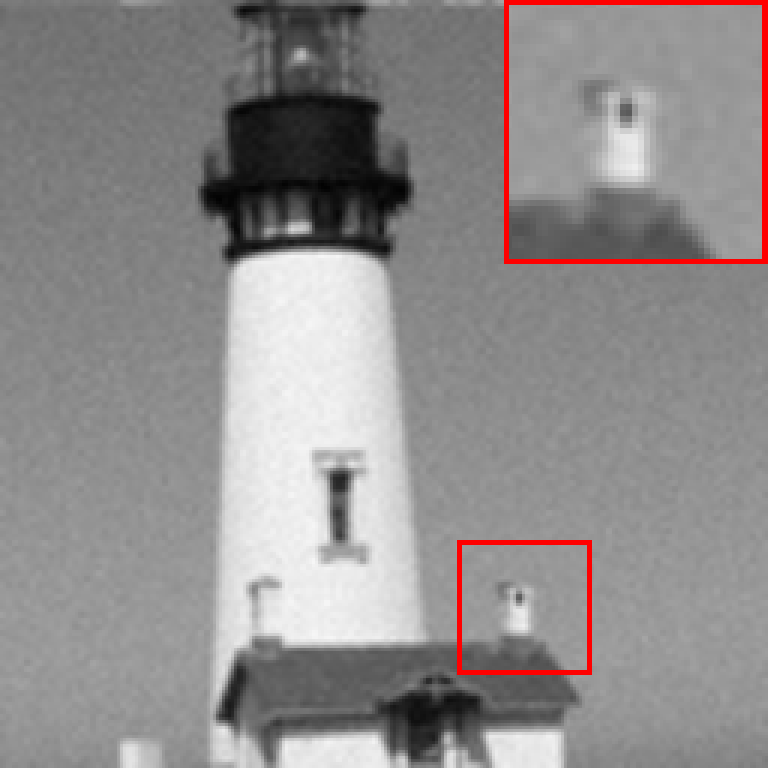}}
  \centerline{(b)}\medskip
\end{minipage}
\begin{minipage}[b]{.19\textwidth}
  \centering
  \centerline{\includegraphics[width=0.98\linewidth]{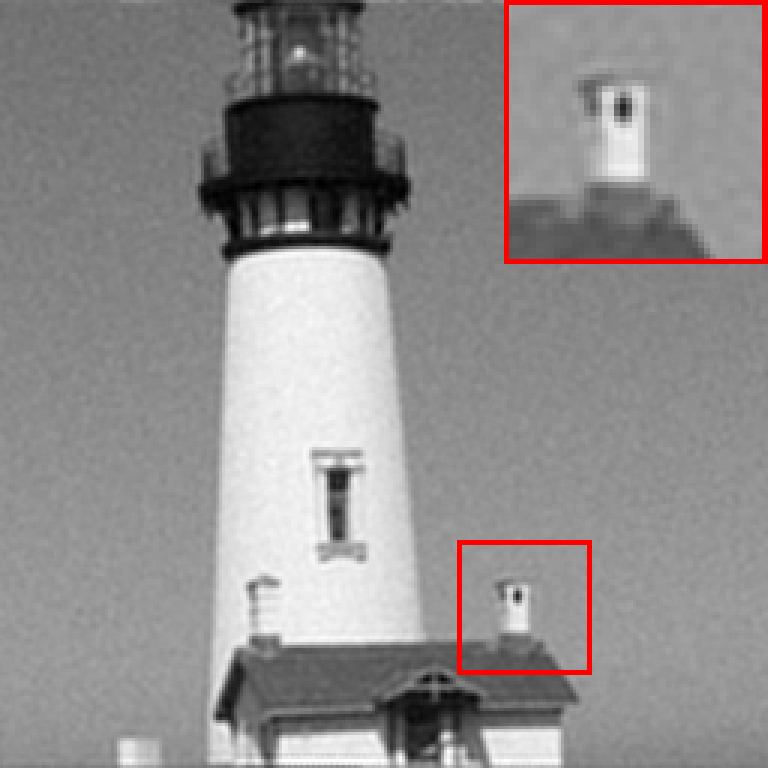}}
  \centerline{(c)}\medskip
\end{minipage}
%
\begin{minipage}[b]{.19\textwidth}
  \centering
  \centerline{\includegraphics[width=0.98\linewidth]{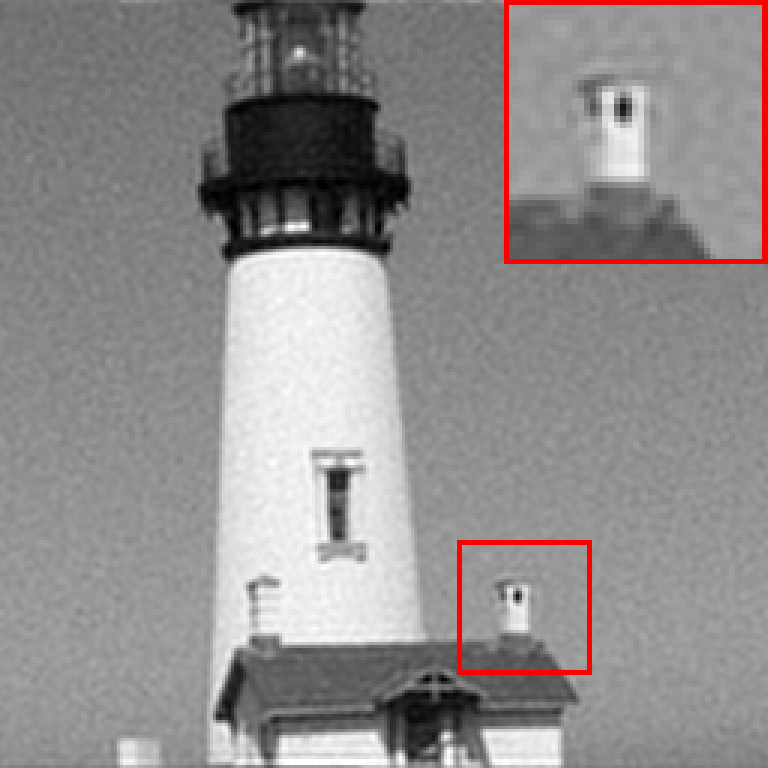}}
  \centerline{(d)}\medskip
\end{minipage}
\begin{minipage}[b]{.19\textwidth}
  \centering
  \centerline{\includegraphics[width=0.98\linewidth]{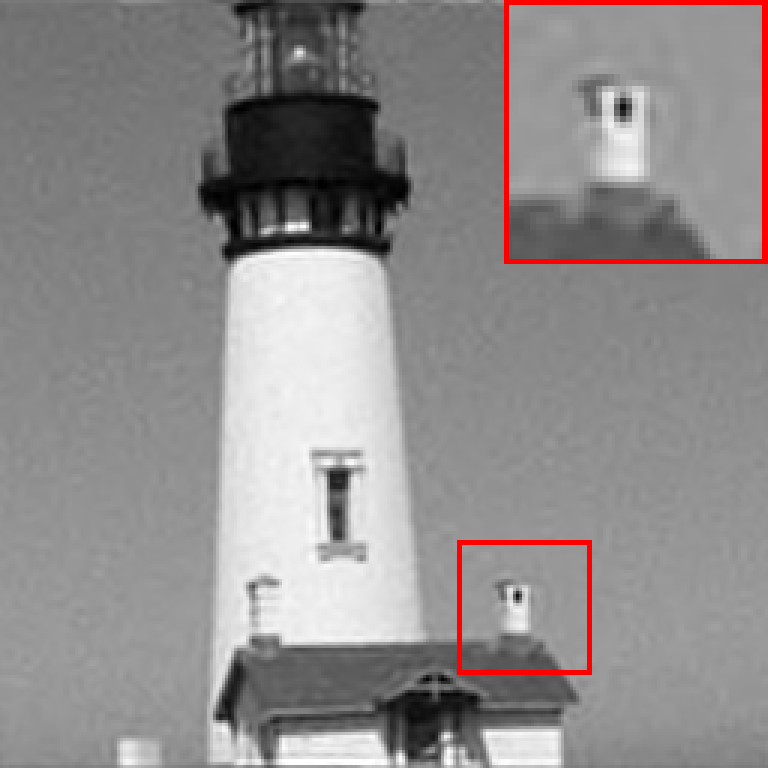}}
  \centerline{(e)}\medskip
\end{minipage}
\vspace{-0.4cm}
\caption{Sample of 200th frame of a reconstructed sequence. (a) Original image. (b) Bicubic interpolation (MSE=18.49dB, SSIM=0.883). (c) LTSR-LMS (MSE=11.80dB, SSIM=0.898). (d) MTSR-LMS (MSE=12.03dB, SSIM=0.8367). (e) WMTSR-LMS (MSE=11.77dB, SSIM=0.934).}
\label{fig:resIdeal}
\end{figure*}

Furthermore, given enough time all SRR methods reach a similar steady-state MSE.
However, the nonlinear Wavelet based regularization of the WMTSR-LMS provided an improved robustness against noise, leading to a better representation of smooth structures without compromising the edges and high frequency structures.
This effect can be most clearly perceived in video sequences that contain significantly smooth regions, such as the one illustrated in Figure~\ref{fig:resIdeal}. In this image, the WMTSR-LMS algorithm is able to provide results with smooth backgrounds largely free from noise without compromising the edges and high frequency structures, unlike the other algorithms.

Note also that the MTSR-LMS algorithm achieves a higher steady state MSE and more noisy result when compared to the gradient-based solution. This happens since the MTSR-LMS is based on only an approximate solution to the cost function in~\eqref{eq:Lagrangian}.
Furthermore, the influence of registration errors, which have a regularizing effect in gradient-based solutions~\cite{borsoi2019SRR_adap_TIP}, might affect the approximate inverse filterbank in~\eqref{eq:filterbank_inv_2}/\eqref{eq:filterbank_inv_3} differently.


\begin{table*} [htb]
\small
\caption{Average PSNR and SSIM for the videos in Example~2.}
\vspace{-0.25cm}
\centering
\renewcommand{\arraystretch}{1.1}
\setlength\tabcolsep{3.5pt}
\resizebox{\linewidth}{!}{%
\begin{tabular}{l|cccccccccc|c}
\hline
 & Boats & Bus & Construction & Conveyor & Kids & Parking & Mall & Market & Street & Train & Mean\\
\hline
Bicubic   & 28.5 / 0.893  &  33.3 / 0.906  &  26.1 / 0.843  &  31.4 / 0.923  &  25.7 / 0.826  &  26.7 / 0.858  &  27.3 / 0.851  &  25.3 / 0.821  &  25.1 / 0.839  &  27.4 / 0.815  &  27.67 / 0.858\\
LTSR-LMS  & 34.0 / 0.935  &  34.9 / 0.916  &  29.0 / 0.889  &  33.8 / 0.926  &  30.0 / 0.914  &  30.6 / 0.909  &  29.6 / 0.894  &  27.2 / 0.865  &  28.4 / 0.904  &  29.8 / 0.877  &  30.73 / 0.903\\
MTSR-LMS  & 34.0 / 0.916  &  35.6 / 0.903  &  28.7 / 0.861  &  34.9 / 0.910  &  29.9 / 0.897  &  30.6 / 0.888  &  30.6 / 0.892  &  28.1 / 0.873  &  29.1 / 0.898  &  30.8 / 0.888  &  31.22 / 0.893\\
WMTSR-LMS & \textbf{34.5} / \textbf{0.947}  &  \textbf{36.5} / \textbf{0.941}  &  29.1 / \textbf{0.905}  &  \textbf{35.6} / \textbf{0.954}  &  30.3 / \textbf{0.922}  &  \textbf{31.0} / \textbf{0.926}  &  30.5 / \textbf{0.909}  &  28.0 / \textbf{0.886}  &  28.9 / \textbf{0.911}  &  30.7 / \textbf{0.895}  &  \textbf{31.51} / \textbf{0.920}\\
CNN       & 33.3 / 0.904  &  35.2 / 0.885  &  \textbf{29.6} / 0.864  &  35.1 / 0.899  &  \textbf{30.5} / 0.896  &  30.6 / 0.880  &  \textbf{31.1} / 0.891  &  \textbf{28.5} / 0.875  &  \textbf{29.7} / 0.901  &  \textbf{31.1} / 0.887  &  31.46 / 0.888\\
Bayesian  & 31.1 / 0.873  &  34.1 / 0.852  &  27.8 / 0.838  &  33.0 / 0.870  &  28.4 / 0.863  &  28.3 / 0.840  &  29.4 / 0.857  &  27.0 / 0.834  &  27.6 / 0.854  &  29.3 / 0.850  &  29.59 / 0.853\\
\hline
\end{tabular}}
\label{tab:algs_PSNR_ex4}
\end{table*}



\begin{figure}[!h]
    \centering
    \includegraphics[width=0.75\linewidth]{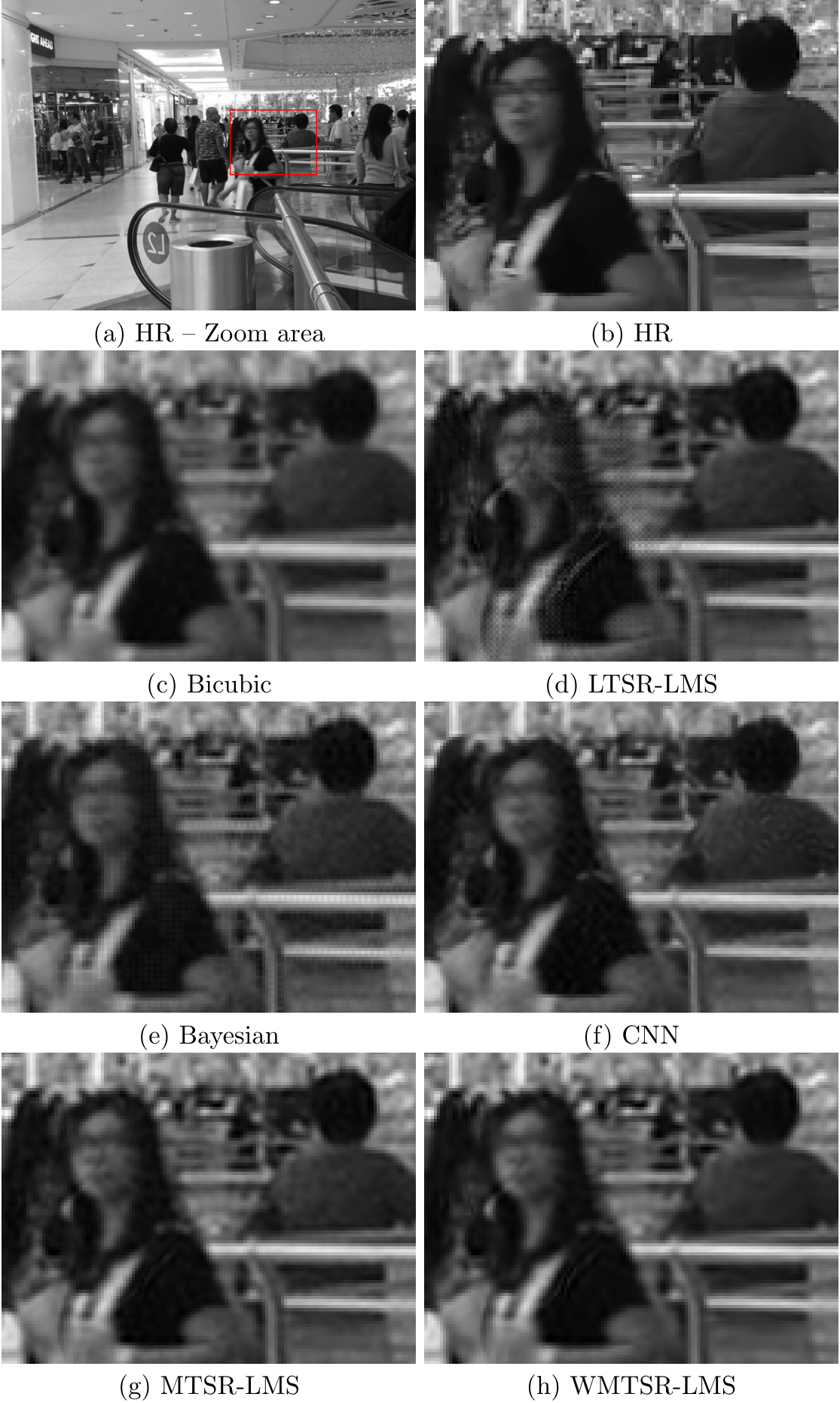}
    \caption{Sample of the 70th frame from the \textit{Mall} video sequence.}
    \label{fig:ex2_visual_video1}
\end{figure}

\begin{figure}[!h]
    \centering
    \includegraphics[width=0.75\linewidth]{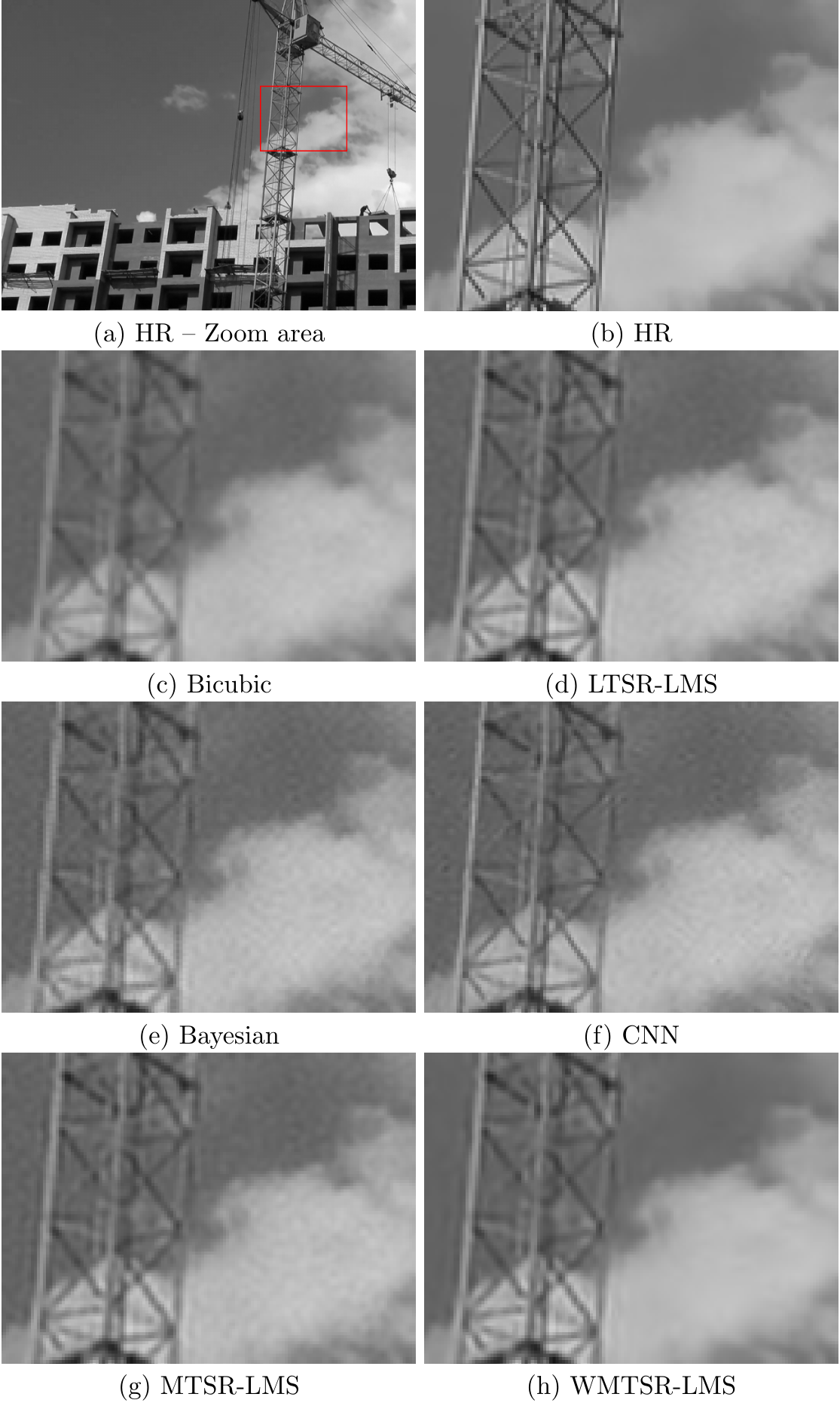}
    \caption{Sample of the 100th frame from the \textit{Construction} video sequence.}
    \label{fig:ex2_visual_video2}
\end{figure}

\begin{figure}[!h]
    \centering
    \includegraphics[width=0.75\linewidth]{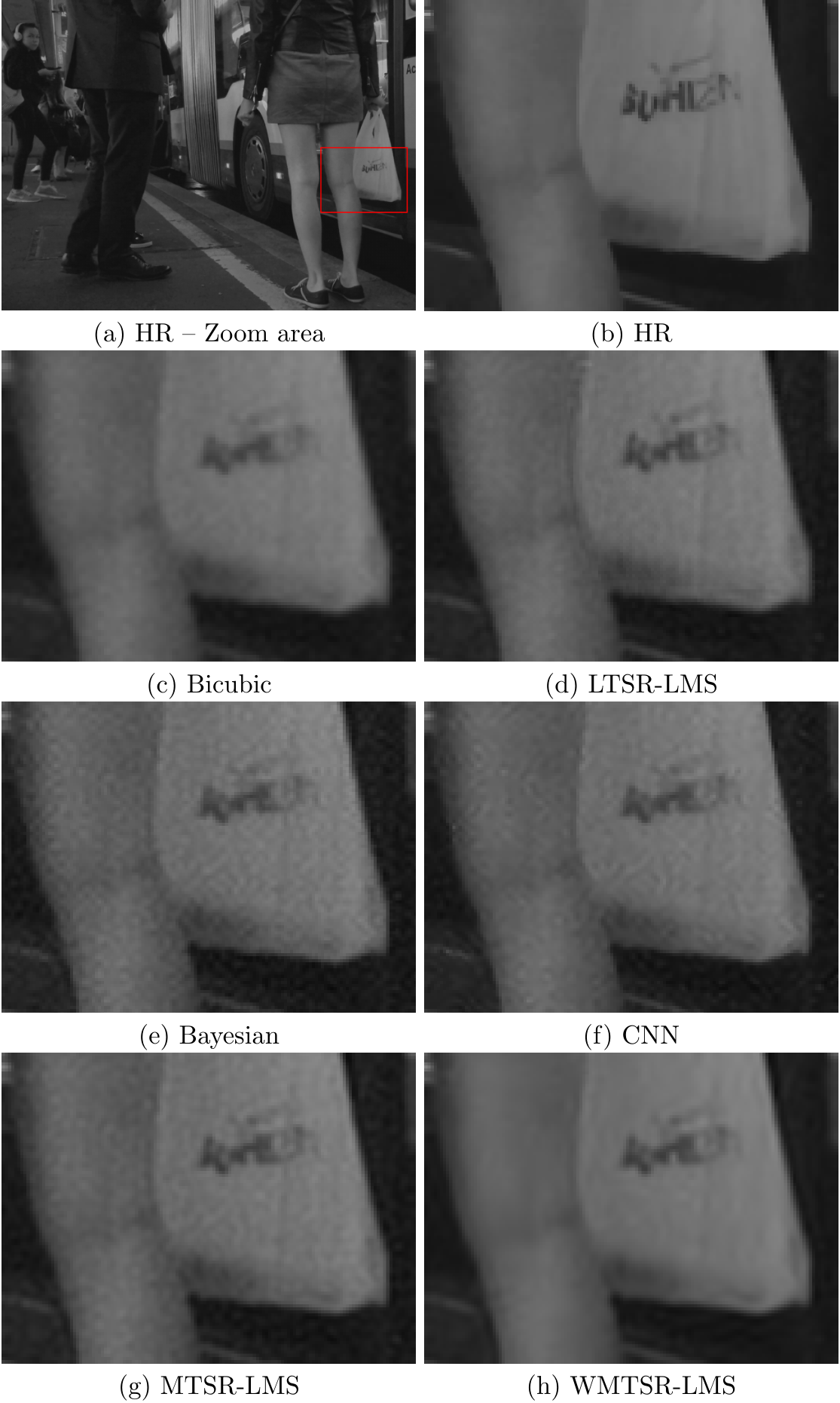}
    \caption{Sample of the 100th frame from the \textit{Bus} video sequence.}
    \label{fig:ex2_visual_video3}
\end{figure}

\subsection{Example 2} \label{sec:examples_real_i}

In this example, we illustrate the performance of the proposed method when super-resolving real video sequences.
To this end, we considered ten high-resolution video sequences extracted from the website \url{videos.pexels.com}. 
To allow for a quantitative evaluation, the original videos were resized to $640\times480$ pixels and used as available HR image sequences, and the degraded LR images were generated from them following the same procedure as in Example~1.

We also compare the performances of the proposed methods to other more recent state-of-the-art algorithms, namely, an adaptive Bayesian method~\cite{liu2014bayesianVideoSRR} and a Convolutional Neural Network (CNN)~\cite{tao2017detailRevealingSRRneuralNets}.
The parameters of the LTSR-LMS, MTSR-LMS and WMTSR-LMS methods were the same as those used in Example~1, and are displayed in Table~\ref{tab:parametersOutlier}.
The \textit{Horn \& Shunck} registration algorithm of~\cite{Sun10} was used in all methods except for the CNN, which had an embedded registration process.
The Bayesian~\cite{liu2014bayesianVideoSRR} and the CNN~\cite{tao2017detailRevealingSRRneuralNets} SRR methods were implemented using codes provided by the respective authors. The simulations were executed on a mobile computer with an Intel Core~2 Duo processor with 2.4Ghz.

The peak signal to noise ratio (PSNR) and structural similarity index (SSIM) for all algorithms and video sequences is shown in Table~\ref{tab:algs_PSNR_ex4}.
It can be seen that the WMTSR-LMS achieved the best average PSNR and SSIM among all tested algorithms.
Moreover, although for some videos sequences the CNN achieved a better PSNR, the WMTSR-LMS resulted in a better SSIM in all cases.
Although the PSNR performance of the CNN and of the MTSR-LMS methods was very similar and close to the PSNR achieved by the WMTSR-LMS, their SSIM was considerably lower. This occurs due to their solutions being considerably noisy, as can be attested through a visual inspection of the results in Figures~\ref{fig:ex2_visual_video1},~\ref{fig:ex2_visual_video2} and~\ref{fig:ex2_visual_video3}.

The gradient-based solution or LTSR-LMS, although achieving a relatively high SSIM and being less affected by noise when compared to the MTSR-LMS, CNN and Bayesian algorithms, suffers from its lack of robustness to innovations. Although the overall difference in PSNR might appear to be small, it reflects significant amounts of artifacts present in regions containing localized motion, which can be seen for instance in a sample of the the Mall video sequence, shown in Figure~\ref{fig:ex2_visual_video1}. In this frame, there is a considerable amount of artifacts in the region comprising the woman's head in the LTSR-LMS result, as opposed to much clearer reconstruction results presented by the MTSR-LMS and WMTSR-LMS algorithms. Furthermore, the Bayesian and especially the CNN methods provide slightly better results in this region, showing robustness to innovations albeit at the expense of a larger computational cost, as will be seen in the next example.

In video sequences not affected by innovation outliers, the WMTSR-LMS algorithm shows a significant performance improvement when compared to the other methods. It is able to reconstruct smooth regions without being significantly affected by noise and without compromising the preservation of sharp image edges.
%
This is illustrated in samples of the Construction and of the Bus video sequences, shown in Figures~\ref{fig:ex2_visual_video2} and~\ref{fig:ex2_visual_video3}, respectively\footnote{Samples from reconstruction results for the remaining video sequences in this example show similar behavior, and can be found in a supplemental material.}. In Figure~\ref{fig:ex2_visual_video2}, the WMTSR-LMS achieves a clear reconstruction of the background without affecting the scaffolding structure. In Figure~\ref{fig:ex2_visual_video3}, the reconstruction of the woman's leg and shopping bag is significantly more accurate than the remaining algorithms, being less affected by artifacts or by residual noise. Even in sequences containing large amounts of innovation outliers, improvements can be observed for the WMTSR-LMS results in the remaining image regions, such as in the man's shirt in the background of Figure~\ref{fig:ex2_visual_video1}.

\begin{figure}[!h]
    \centering
    \centerline{\includegraphics[width=\linewidth]{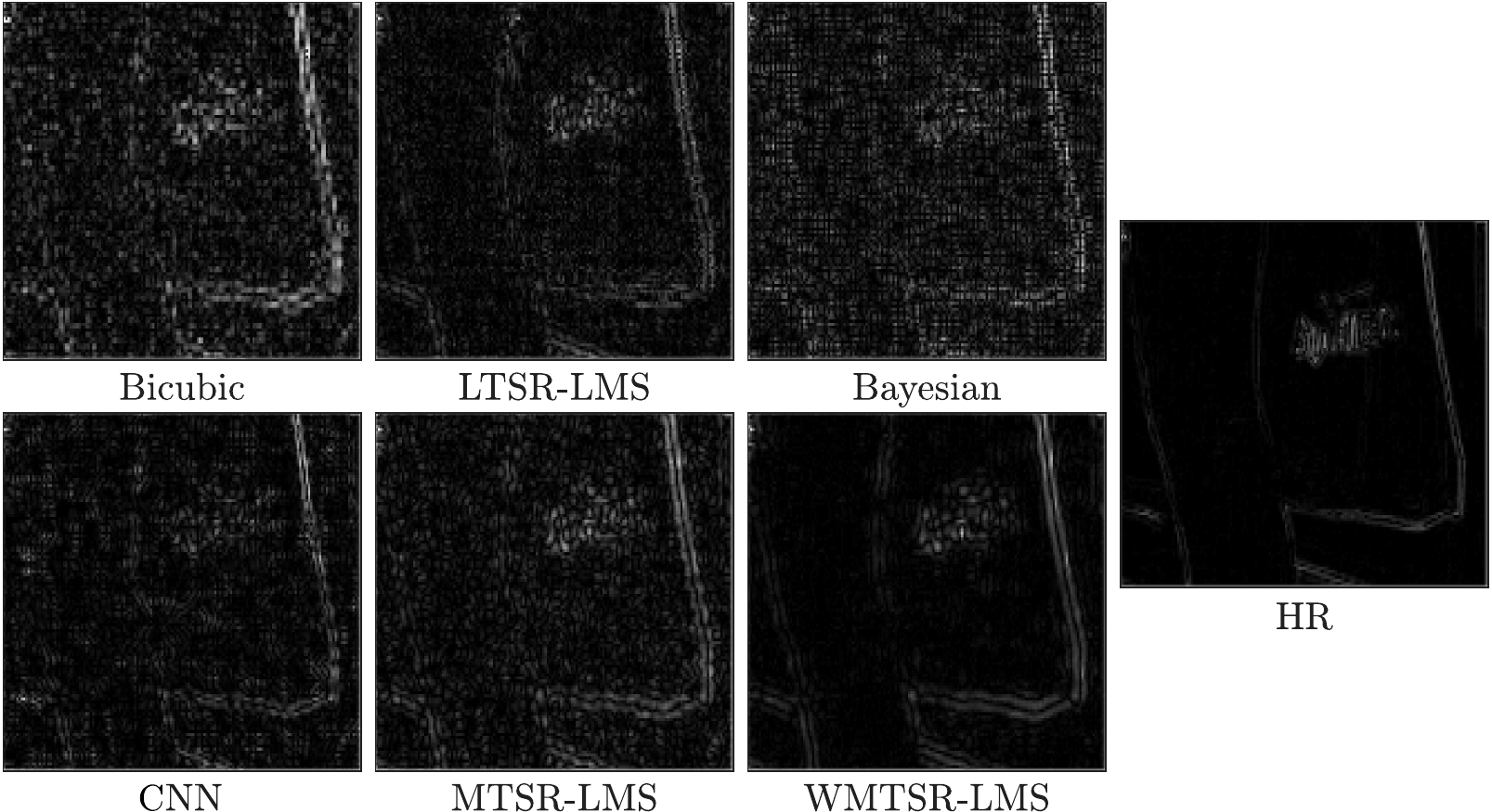}}
    \caption{Edge map of the zoomed region of the 100th frame from the \textit{Bus} video sequence in Figure~\ref{fig:ex2_visual_video3}.}
    \label{fig:edges_bus}
\end{figure}

To see how the improved robustness of the WMTSR-LMS algorithm does not significantly affect sharp image edges in the reconstructions, we can look at the edge maps of the solutions from the \textit{Bus} video sequence, which are shown in Figure~\ref{fig:edges_bus}. It can be seen that the image edges of the solution obtained by the WMTSR-LMS algorithm closely agree with those of the HR image. Moreover, even though homogeneous image regions are smoother when compared to the reconstructions obtained by the other algorithms, sharp edges are still well preserved and follow closely those in the HR edge map.


The execution times of all algorithms is presented in Table~\ref{tab:algs_processing_time}. It can be seen that the MTSR-LMS algorithm incurred in a processing time that is very similar to the gradient based solution. The WMTSR-LMS, on the other hand, required about twice the time taken by the MTSR-LMS, due to the edge-preserving regularization. Nevertheless, the WMTSR-LMS processing time it is still orders of magnitude smaller than that required by the remaining SRR methods (i.e. Bayesian and the CNN), despite showing similar or better reconstruction quality.



\begin{table} [htb]
\scriptsize
\caption{Average processing time (in seconds) per frame for the videos in Example~2.}
\vspace{-0.35cm}
\begin{center}
\renewcommand{\arraystretch}{1.1}
\setlength\tabcolsep{3.5pt}
{%
\begin{tabular}{ccccccc}
\hline
Bicubic & LTSR-LMS & MTSR-LMS & WMTSR-LMS & Bayesian~\cite{liu2014bayesianVideoSRR} & CNN~\cite{tao2017detailRevealingSRRneuralNets} \\
\hline
0.012 & 0.683 & 0.795 & 1.957 & 66.38 & 83.54 \\
\hline
\end{tabular}}
\end{center}
\label{tab:algs_processing_time}
\end{table}

\section{Conclusions} \label{sec:concl}

In this paper, a new low-complexity adaptive video super-resolution reconstruction algorithm was proposed with improved robustness to innovations and to additive noise.
We use a nonlinear cost function constructed considering the characteristics of typical innovation outliers in natural image sequences and an edge-preserving regularization strategy.
This cost function is optimized using a specific alternating projections strategy over non-convex sets that is able to converge in very few iterations.
Furthermore, benefiting from multidimensional multirate signal processing tools, an accurate approximate solution to each projection operation can be computed very efficiently. This solves the slow convergence issue of stochastic gradient-based methods while keeping a small computational complexity, making the algorithm suitable to real-time processing applications.
Simulation results with both synthetic and real video sequences show that the proposed algorithm performs similarly or better than state of the art super-resolution methods, with only a small fraction of their computational complexity.




\bibliographystyle{IEEEtran}
\bibliography{references_fbank}

\end{document}